\documentclass{article}

\PassOptionsToPackage{round}{natbib}

\usepackage[main,final]{neurips_2026}

\usepackage[american]{babel}                     %
\usepackage{mathtools}                           %
\usepackage{tikz}                                %
\usepackage[utf8]{inputenc}                      
\usepackage{natbib}                              
\usepackage[T1]{fontenc}                         
\usepackage{hyperref}                            
\usepackage{cleveref}
\usepackage{url}                                 
\usepackage{booktabs}                            
\usepackage{amsthm,amsmath,amssymb,amsfonts}     
\usepackage{thmtools,thm-restate}                
\usepackage{mdframed}                            
\usepackage[many]{tcolorbox}                     
\usepackage{tabularx}                            
\usepackage{nicefrac}                            
\usepackage{microtype}                           
\usepackage{xcolor}                              
\usepackage{bm}                                  
\usepackage{graphicx}                            
\usepackage{enumitem}                            
\usepackage{aliascnt}                            
\usepackage{sidecap}                             
\usepackage{makecell}                            
\usepackage{array}                               
\usepackage{subcaption}                          %
\usepackage{longtable}                           

\declaretheoremstyle[
  spaceabove=5pt,
  spacebelow=5pt,
  headfont=\normalfont\bfseries,
  notefont=\mdseries,
  notebraces={(}{)},
  bodyfont=\normalfont,
  postheadspace=1em,
  qed=,
  headpunct={},
]{theoremstyle}

\declaretheorem[name=Theorem,style=theoremstyle,numberwithin=section]{theorem}
\declaretheorem[name=Proposition,style=theoremstyle,numberwithin=section]{proposition}

\declaretheorem[name=Assumption,style=theoremstyle,numberwithin=section]{assumption}

\def\R{\mathbb R}

\def\Pc{\mathcal P}

\newcommand{\rmd}{\mathrm{d}}

\newcommand{\eqsp}{\,}
\newcommand{\ola}{\overleftarrow}

\newcommand{\Id}{\mathbf{I}}

\newcommand{\pidata}{\pi_{\mathrm{data}}}
\def\eqlaw{\stackrel{\mathcal L}{=}}
\NewDocumentCommand{\csp}{m m o}{\IfValueTF{#3}{\mathcal{C}^{#1}\left(#2; #3\right)}{\mathcal{C}^{#1}\left(#2\right)}}

\newcommand{\normEc}[1]{\left\|#1\right\|}

\newcommand{\normFr}[1]{\left\|#1\right\|_{\operatorname{F}}}

\NewDocumentCommand{\wasserstein}{m o o}{
  \IfValueTF{#3}{\mathcal{W}_{#1}\left(#2, #3\right)}
  {\mathcal{W}_{#1}}}
\newcommand{\brownvar}[1]{\operatorname{B}_{#1}}
\newcommand{\tildebrownvar}[1]{\widetilde{\operatorname{B}}_{#1}}
\newcommand{\dbrown}[1]{\rmd\!\operatorname{B}_{#1}}
\newcommand{\dtildebrown}[1]{\rmd\widetilde{\operatorname{B}}_{#1}}

\def\st{\emph{s.t.},~}
\def\eg{\emph{e.g.},~}
\def\ie{\emph{i.e.},~}

\definecolor{tab:blue}{RGB}{31, 119, 180}
\definecolor{tab:orange}{RGB}{255, 127, 14}
\definecolor{linkcolor}{RGB}{44, 96, 163}
\hypersetup{
    colorlinks=true,
    citecolor=linkcolor,
    linkcolor=linkcolor,
    urlcolor=linkcolor,
}

\title{Do Heavy Tails Help Diffusion? On the Subtle Trade-off Between Initialization and Training}

\author{%
  Hamza Cherkaoui\textsuperscript{1}\thanks{Equal contribution.} \quad
  H\'el\`ene Halconruy\textsuperscript{1,2}\footnotemark[1] \quad
  Antonio Ocello\textsuperscript{3} \\[0.75em]
  \textsuperscript{1} SAMOVAR, T\'el\'ecom SudParis, Institut Polytechnique de Paris, Palaiseau, France\\
  \textsuperscript{2} Modal'X, Universit\'e Paris Nanterre, Nanterre, France\\
  \textsuperscript{3} CREST, CNRS, Ecole polytechnique, Groupe ENSAE-ENSAI, ENSAE Paris,\\
  Institut Polytechnique de Paris, 91120 Palaiseau, France\\
  \texttt{\{hamza.cherkaoui,helene.halconruy\}@telecom-sudparis.eu}\\
  \texttt{antonio.ocello@ensae.fr} \\
}

\begin{document}

\maketitle

\begin{abstract}
Recent works have proposed incorporating heavy-tailed (HT) noise into diffusion- and flow-based generative models, with the goals of better recovering the tails of target distributions and improving generative diversity.
This motivation is intuitive: if the data are heavy-tailed, HT noise may appear better matched than light-tailed (LT) Gaussian noise.
However, replacing Gaussian noise by HT noise also changes the underlying estimation problem.
In this paper, we revisit this paradigm through a combined theoretical and empirical study, establishing sampling-error bounds for two representative diffusion models driven by HT and LT noise.
We show that HT diffusion models can achieve exponentially fast initialization, but accumulated training errors may offset this advantage.
Experiments on synthetic and real-world datasets support our theoretical findings.
Our results call into question a growing design trend in generative modeling and challenge the use of HT noise to improve rare-region exploration.
\end{abstract}


\section{Introduction}
\label{sec:intro}

\emph{Do heavy tails help?}
At first glance, the answer seems obvious.
When the target distribution exhibits heavy-tailed (HT) behavior, replacing Gaussian noise with HT alternatives in diffusion models appears both natural and theoretically justified.
Indeed, a growing body of literature adopts this principle, aiming to improve tail recovery by aligning the model with the data.
In this work, we challenge this paradigm.
We show that HT diffusion models can achieve exponentially fast initialization, but training errors accumulate along the reverse chain and may outweigh this advantage in finite-sample regimes.

Modern generative methods, such as diffusion and flow-based approaches, can be viewed as mechanisms that transport a simple reference distribution, typically Gaussian, into a complex target distribution through a continuous-time transformation.
In diffusion models \citep[see, \eg][]{sohl2015deep,song2019generative,ho2020denoising,song2021denoising,song2020score}, this is achieved by defining a forward noising process that progressively corrupts the data, and a reverse-time dynamics driven by a learned score function that reconstructs samples. Flow-based models \citep{lipman2023flowmatching,boffi2023probability,albergo2025stochastic,liu2023flowstraight} instead parameterize this transport through continuous-time dynamics governed by neural vector fields.
Despite their apparent differences, both frameworks rely on accurately estimating the score or vector field associated with intermediate distributions along the path.
This estimation problem is central: the quality of the learned model ultimately depends on how well these quantities can be approximated from finite data.

Such approaches have rapidly gained traction across a wide range of disciplines, including finance \citep{sattarov2023findiff,guo2025diffusion}, weather forecasting \citep{landry2026generating,couairon2026archesweathergen}, and medical imaging \citep{bedin2025reconstructing}.
In these applications, the goal is not only to generate high-quality samples but also to faithfully represent the full complexity of the underlying data distribution, especially rare and extreme events, ranging from catastrophic risks in finance and insurance to fine-grained and underrepresented structures in high-dimensional data.
This is especially important in domains such as meteorology, insurance, and risk modeling, where rare or extreme events play a central role.
Failing to model such regions can lead to poor risk assessment, lack of robustness, or loss of diversity.

Representing the tails of a distribution poses a fundamental challenge: by construction, these regions are sparsely observed in finite datasets, making them difficult to learn.
As a result, the ability to recover such regimes depends critically on how well the underlying learning procedure extrapolates beyond high-density regions.
This has naturally motivated recent efforts to design generative mechanisms that better explore and capture the tails of the data distribution during both training and sampling.

 \paragraph{Heavy-tailed extensions of diffusion and flow models.}
Recent work has proposed replacing the Gaussian ingredients of diffusion and flow-based generative models, either in the reference distribution or the noising process, to better accommodate \emph{heavy-tailed target distributions}.
Among these contributions, Denoising L\'evy Probabilistic Models (DLPM) replace the Gaussian increments of DDPM \citep{ho2020denoising} with isotropic $\alpha$-stable noise and adapt the denoising objective to learn the reverse dynamics \citep{shariatian2025heavy}.
We refer to~\autoref{sec:supp-related-work} for a detailed discussion of related approaches.
In this paper, we revisit this paradigm from a statistical estimation perspective, examining whether matching the noise to the tails of the data translates into improved learning and sampling performance.

\paragraph{Contributions.}
At a high level, these contributions reflect the growing interest in HT generative modeling, but they also leave open a broader question:
\begin{center}
     \emph{Do heavy tails help, or do they make learning harder?}
\end{center}
\noindent\textbf{\textit{A twofold approach.}}
Since existing approaches are often studied under different assumptions, metrics, and experimental settings, their respective strengths and limitations remain poorly understood.
In particular, it is still unclear whether HT mechanisms genuinely improve sampling performance once the statistical difficulty of learning the reverse dynamics is taken into account.
We address this question from two \textit{complementary} perspectives:

$\bullet$ \emph{From a theoretical viewpoint}, we compare Gaussian variance-exploding (VE) diffusion and DLPM for generating $\alpha$-stable distributions.
We derive non-asymptotic bounds in total variation distance, decomposing sampling error into initialization and training errors.
Our analysis reveals a fundamental \textit{trade-off}.
While DLPM achieves substantially smaller initialization error due to its forward dynamics rather than heavy tails, this gain may be offset by larger training errors.
In particular, approximation and estimation errors accumulate over the full horizon $T$ for DLPM, whereas the Gaussian VE diffusion bound is primarily controlled by the early-stopping parameter $t_0$.
This suggests that the statistical difficulty of learning the reverse dynamics may dominate the initialization advantage of HT models in finite-sample regimes.

$\bullet$ \emph{From an empirical viewpoint}, we compare DLPM against DDPM, DDIM, t-EDM, and linear Gaussian flow matching.
We assess global distributional fit using MMD$^2$ and tail coverage using the Tail Coverage Error (TCE), defined later in the paper.
We also investigate rare-class recovery on imbalanced image datasets to examine whether HT diffusion models provide benefits in this setting, as suggested by \citet{shariatian2025heavy}.
Finally, sample-size learning curves and zero-network controls help distinguish the effects of training data availability from those of the sampling dynamics themselves.

\noindent\textbf{\textit{Main insight.}}
Overall, our theoretical and empirical findings challenge the presumed advantage of HT noising for HT generation.
Although HT initialization reduces initialization error, this gain can be outweighed by the difficulty of learning the reverse dynamics, particularly with limited training data.
Our results show that \textbf{tail alignment does not guarantee better generation}, and that LT alternatives can outperform HT models.
\section{Sampling error decomposition}
\label{sec:theory}

We now introduce a sampling-error decomposition that isolates the main sources of discrepancy between the generated distribution and the target.
Specifically, we decompose the global error into two main contributions: an \textbf{initialization error} and a \textbf{training error}, the latter further splitting into \textbf{approximation} and \textbf{finite-sample estimation} components , with non-asymptotic bounds for each term.

This decomposition serves two purposes.
First, it yields a \emph{transparent interpretation} of how different components of the pipeline contribute to the overall error.
Second, it enables a \emph{quantitative comparison} between two representative formulations: a variance-exploding Gaussian diffusion (VE-GD) on the LT side (see \autoref{subsec:theory_lighttail}), and DLPM on the HT side.
To make this comparison concrete, we measure discrepancies in \textbf{total variation (TV) distance}, which provides a strong, distribution-level notion of error: it directly controls the worst-case difference in expectations over bounded observables and is sensitive to mass in the tails, making it particularly well-suited to distinguish HT and LT behaviors.

\paragraph{Notations.}
We denote by $\Id_d$ the $d\times d$ identity matrix and by $\Pc\left(\R^d\right)$ the set of probability distributions $\mu$ on $\R^d$.
We write $X\eqlaw Y$ when the random variables $X$ and $Y$ have the same distribution.
We denote by $\csp{k}{\R^{d_0}}[\R^{d_1}]$ the space of $k$-times continuously differentiable functions from $\R^{d_0}$ to $\R^{d_1}$ and by $W^{\beta,2}(\R^{d_0})$ the Sobolev space of functions on $\mathbb R^d$ possessing square-integrable weak derivatives up to order $\beta$ (in the fractional sense when $\beta\notin\mathbb N$).
When $d_1=1$, we simplify the notation to $\csp{k}{\R^{d_0}}$. 
We write $a \lesssim b$ if there exists a constant $C>0$ such that $a\le Cb$, and $a\asymp b$ if $a\lesssim b$ and $b\lesssim a$.
When there is no ambiguity, $\normEc{\cdot}$ denotes the Euclidean norm on $\R^d$, and $\normFr{\cdot}$ denotes the Frobenius norm for matrices.
We denote by $\Pc(\R^d)\ni\mu\mapsto\widehat{\mu}$ the Fourier transform of a function or distribution \citep[see, \eg][]{folland1999real}.

\subsection{Sampling error of VE-GD}
\label{subsec:theory_lighttail}

\paragraph{Gaussian forward dynamics and terminology.}
Gaussian score-based diffusion models \citep{song2020score} can be built on different forward dynamics. We consider the \emph{variance-exploding} (VE) formulation, in which Gaussian noise is added to the data without rescaling, so that the noise variance grows with time.
Our analysis relies on \citet{yu2026diffusionheavytailedtargets}, who provide explicit statistical guarantees for score estimation and sampling with heavy-tailed targets for a VE diffusion with constant diffusion coefficient; we therefore adopt this setting: the forward process $(X_t)_{t\in[0,T]}$ solves
\begin{equation}
    \label{eq:general-fwd-dynamics}
    \rmd X_t = \dbrown{t}\eqsp,
    \qquad X_0 \sim \pidata\eqsp,
\end{equation}
with $(\brownvar{t})_{t\in[0,T]}$ a standard Brownian motion.
We refer to this Gaussian VE model as\emph{VE-GD}; all statements below refer to the dynamics \autoref{eq:general-fwd-dynamics}.
By linearity, with $Z\sim \mathcal N(0,I_d)$ independent of $X_0$ and $\phi_t$ the density of $\mathcal N(0,t I_d)$,
\begin{equation}
    \label{eq:DDPM_convolution}
    X_t \eqlaw X_0 + \sqrt{t}\,Z
    \eqsp,
    \qquad
    p_t = \pidata * \phi_t\eqsp.
\end{equation}

Following the classical theory of diffusion time reversal \citep{anderson1982reverse,haussmann1986time,cattiaux2023time}, there exists a process $(\ola X_t)_{t\in[0,T]}$ such that $\mathcal{L}\big((X_t)_{t\in[0,T]}\big) = \mathcal{L}\big((\ola X_{T-t})_{t\in[0,T]}\big)$.
A fundamental result of \citet{haussmann1986time} shows that, under mild regularity assumptions, this time-reversed process is itself governed by an SDE whose drift depends explicitly on the score function of the forward marginals, namely $(t,x)\mapsto \nabla \log p_t(x)$.
This characterization forms the theoretical basis of score-based generative methods, as it allows the reverse dynamics to be simulated through score estimation.
In the present setting, the reverse-time dynamics associated with \autoref{eq:general-fwd-dynamics} takes the form
\begin{equation*}
    \mathrm d \ola X_t = \nabla \log p_{T-t}(\overleftarrow X_t)\,\mathrm dt + \dtildebrown{t} \eqsp,
\end{equation*}
with $(\tildebrownvar{t})_{t\ge0}$ a Brownian motion.

\paragraph{HT assumption.}
Throughout the paper, we assume that the target distribution satisfies the following polynomial tail condition.
\begin{assumption}[Tail behavior]
\label{ass:tail-behavior}
    The target density $\pidata$ has polynomial tail index $\gamma>1$, \ie
    \begin{equation*}
        \mathbb P_{X_0\sim\pidata}(|X_0|>r)\sim r^{-\gamma}
        \quad
        \text{ as }r\to\infty
        \eqsp.
    \end{equation*}
\end{assumption}
\vspace{-1em}

The Gaussian smoothing in \autoref{eq:DDPM_convolution} regularizes the density at small scales while preserving polynomial tail behavior: if $\pidata$ has tail index $\gamma$, then typically \citep[see, \eg Theorem
1.2 in][]{nolan2020univariate} we also have
\begin{align*}
    \mathbb P(|X_t|>r)\sim r^{-\gamma}
    \quad
    \text{ as }r\to\infty
    \eqsp.
\end{align*}

\paragraph{Score approximation error.}
In practice, the score function $\nabla \log p_t$ is learned from data using neural network parameterizations $s_\theta(x,t)$ trained via score matching objectives \citep{hyvarinen2005estimation,vincent2011connection,efron2011tweedie}.
This approximation induces a statistical error that propagates along the reverse-time dynamics.
To quantify both the score estimation error and the bias introduced by Gaussian smoothing, we rely on the framework of \citet{yu2026diffusionheavytailedtargets}.

Beyond the tail assumptions described above, where the parameter $\gamma$ governs the tail behavior of the distribution, we need a regularity condition that controls its smoothness at small scales.
Suppose that the target density $\pidata$ satisfies the following Sobolev regularity condition.
\begin{assumption}\label{ass:DDPM_regularity}
    For some $\beta\in(0,2]$ and $L>0$,
    \begin{equation*}
        \pidata \in W^{\beta,2}(\mathbb R^d)\eqsp,
        \qquad
        \int_{\mathbb R^d}
        (1+|\xi|^2)^{\beta}
        |\widehat \pidata(\xi)|^2\,\rmd\xi \le L^2\eqsp.
    \end{equation*}
\end{assumption}
\vspace{-1.5em}

\paragraph{Sampling error decomposition for VE-GD.}
Under the previous tail and regularity assumptions, the following theorem quantifies the different sources of error arising in the generative process of score-based models.
More precisely, it makes explicit the respective contributions of initialization, approximation, and statistical estimation, and provides non-asymptotic bounds for each component in total variation distance.

\begin{theorem}[Sampling error decomposition for VE-GD]
    \label{thm:error-DDPM}
    Suppose that \autoref{ass:tail-behavior} and \autoref{ass:DDPM_regularity} hold.
    Let $(p_t)_{t\in[0,T]}$ be the VE-GD forward marginals defined by \autoref{eq:DDPM_convolution}.
    Denote $\widehat Y$ the practical reverse process initialized from $\phi_T=\mathcal N(0,TI_d)$, driven by a learned score $s_\theta$.
    Then, for any early stopping time $t_0\in(0,T)$, the sampling error satisfies
    \begin{equation}\label{eq:DDPM_sampling_bound}
        \mathrm{TV} \big( \pidata, \mathcal L(\widehat Y_{T-t_0}) \big) \lesssim \underbrace{t_0^{\frac{\beta(\gamma+1)}{d+2(\gamma+1)+2\beta}}
        + \operatorname{polylog}(n)\, n^{-\frac{\gamma+1}{2(d+\gamma+1)}}
        \, t_0^{-\frac{d(\gamma+1)}{4(d+\gamma+1)}}}_{\text{training error (approximation+estimation)}}
        + \underbrace{T^{-1/2}}_{\text{initialization error}} \eqsp.
    \end{equation}
\end{theorem}
The three terms in the bound admit a clear interpretation.
The \emph{first term} corresponds to the approximation error induced by Gaussian smoothing and early stopping, and reflects a bias-variance trade-off governed jointly by the Sobolev regularity $\beta$ and the tail index $\gamma$.
The \emph{second term} captures the estimation error, which depends on the sample size $n$ and deteriorates as $t_0$ decreases, due to the increasing difficulty of estimating the score at small noise levels.
Finally, the \emph{third term} is the initialization error, which arises from the mismatch between the true terminal distribution and the Gaussian prior, and decreases at a polynomial rate in the diffusion horizon $T$. This rate is a property of the VE dynamics \autoref{eq:general-fwd-dynamics}, in which $X_0$ is never rescaled and $p_T$ approaches $\phi_T$ only through the growth of the noise variance.

The training contribution can be written in the generic form
\begin{equation*}
    t_0^{a} + \operatorname{polylog}(n)\, n^{-c}\, t_0^{-b} \eqsp,
\end{equation*}
where
\begin{equation*}
    a = \frac{\beta(\gamma+1)}{d+2(\gamma+1)+2\beta} \eqsp,
    \qquad
    b = \frac{d(\gamma+1)} {4(d+\gamma+1)} \eqsp,
    \qquad
    c = \frac{\gamma+1} {2(d+\gamma+1)} \eqsp.
\end{equation*}
In this decomposition, the first quantity corresponds to the approximation error (also interpretable as an early-stopping bias) while the second one captures the statistical estimation error arising from score learning.
Together, these two quantities exhibit the classical \emph{bias-variance tradeoff}: choosing smaller values of $t_0$ reduces the approximation bias but simultaneously increases the statistical error due to the singular behavior of the score function as $t\to0$.

Optimizing the tradeoff between these two contributions leads to the following optimal early-stopping scale: $t_0\asymp n^{-\frac{c}{a+b}} \eqsp,$ giving the optimized training rate
\begin{equation*}
    \operatorname{polylog}(n)\, n^{-\frac{ca}{a+b}} =
    \operatorname{polylog}(n)\, n^{-\frac{2\beta(\gamma+1)} {4\beta(\gamma+1)+6\beta d+2d(\gamma+1)+d^2}} \eqsp.
\end{equation*}

Hence, the VE-GD training error follows a nonparametric convergence rate in the sense of classical minimax nonparametric estimation theory; see, \eg \cite{Tsybakov2009}.
As expected, the rate deteriorates with the ambient dimension $d$, while it improves with the Sobolev smoothness parameter $\beta$.
Moreover, the dependence on the tail parameter $\gamma$ shows that heavier tails lead to slower statistical convergence: larger $\gamma$ (lighter tails) improve the rate, whereas small $\gamma$ significantly degrade it.
This reflects the additional difficulty of estimating score functions for heavy-tailed targets near small diffusion times.

\paragraph{Structural lower bounds.}
Although a matching minimax lower bound for the overall sampling error appears out of reach, each component of \autoref{eq:DDPM_sampling_bound} is unavoidable on an appropriate subclass of problems.
\emph{\underline{Initialization:}} This term is deterministic and does not depend on $n$: it is a structural mismatch between the terminal marginal $p_T$ and the prior $\phi_T$, which cannot be reduced by collecting more data.
For a translated target $\pidata=\delta_{\mu}*\pi$, the discrepancy $\mathrm{TV}(p_T,\phi_T)$ is of order $\normEc{\mu}\,T^{-1/2}$, so the polynomial rate $T^{-1/2}$ cannot be improved uniformly over the class within the VE dynamics.
\emph{\underline{Approximation:}} One can construct target distributions whose Gaussian-smoothing bias at time $t_0$ is of the same order as the early-stopping term $t_0^{a}$.
\emph{\underline{Estimation:}} Minimax lower bounds for score estimation of heavy-tailed targets are available in \citet{yu2026diffusionheavytailedtargets}.
These lower bounds hold term by term; they do not yield a lower bound for the sum of the three contributions.

\subsection{Sampling error of DLPM}
\label{subsec:theory_heavytail}

\paragraph{$\alpha$-stable distributions.}
A natural way to adapt diffusion models to the HT setting consists in replacing the Gaussian noising process by stable distributions.
This class of distributions arises as the limiting laws in the generalized central limit theorem \citep{gnedenko1968limit}.
In dimension one, an $\alpha$-stable random variable is characterized by a tail index $\alpha\in(0,2]$, together with location, scale, and skewness parameters.
The case $\alpha=2$ corresponds to the Gaussian distribution, whereas $\alpha<2$ yields heavy-tailed laws satisfying
\[
    \mathbb P(|X|>r)\sim r^{-\alpha}
    \qquad \text{as } r\to\infty\eqsp,
\]
and therefore $\mathbb E[|X|^p]<\infty$ if and only if $p<\alpha$ \citep{nolan2020univariate}.
This tail behavior is directly connected to \autoref{ass:tail-behavior} of $\pidata$: the underlying intuition is that $\alpha$ should be comparable to the tail index $\gamma$ of $\pidata$, so that the noising process and the target share a similar tail heaviness.
In particular, smaller values of $\alpha$ correspond to heavier tails and weaker moment properties.

In the multivariate setting, several extensions are possible.
We focus on isotropic, or rotationally invariant, $\alpha$-stable distributions, whose characteristic function is given by
\[
    \mathbb E\!\left[\exp(iu^\top X)\right]
    =
    \exp\!\left(i\mu^\top u-\sigma^\alpha \|u\|^\alpha\right)\eqsp,
    \qquad u\in\mathbb R^d \eqsp.
\]
We denote this law by $S^{\bf i}_\alpha(\mu,\sigma \Id_d)$.
For $\alpha=2$, it coincides with a multivariate Gaussian, while for $\alpha<2$ it provides a canonical heavy-tailed analogue used in stable-noise generative processes.

\paragraph{DLPM.}
We now turn to DLPM, taken as the canonical representative of generative models in the HT setting.
In contrast with VE-GD, the forward noising process is driven by isotropic $\alpha$-stable Lévy increments.
We denote these increments by $A=(A_1,\dots,A_T)$, where each $A_t$ is $\alpha$-stable. Conditionally on $A=a$, the forward path law is denoted by $k_{0:T\mid a}^{(\alpha)}$.

The learned reverse dynamics is modeled as a time-inhomogeneous Markov chain:
\begin{equation}\label{eq:DLPM_Markov_kernel}
    \overleftarrow q_{0:T\mid a}^{\theta} = \overleftarrow q_{T}^{\theta} \prod_{t=1}^T \overleftarrow q_{t-1\mid t,a}^{\theta}\eqsp.
\end{equation}
Exploiting the additive and scaling properties of $\alpha$-stable laws, the forward recursion admits an explicit affine representation.
In particular, the terminal variable can be written as
\begin{equation}
    \label{eq:DLPM_YT}
    Y_T = a_T Y_0 + b_T S_\alpha\eqsp,
\end{equation}
where $S_\alpha\sim S^{\bf i}_\alpha(0, \Id_d)$ is independent of $Y_0$, and $(a_T,b_T)$ are deterministic coefficients determined by the noise schedule.

\paragraph{Sampling error bound for DLPM.}
To analyze the sampling error, we impose regularity conditions on the reverse object and on the forward signal-to-noise ratio.

\begin{assumption}[Hölder regularity of the reverse object]
    \label{ass:holder_reverse}
    Let $s_\alpha^\star:[0,T]\times\mathbb R^d\to\mathbb R^d$ be the true reverse object.
    We assume that there exists $\beta(\alpha)>0$ and a constant $C_{\alpha,\beta}<\infty$ such that, for all $t\in[0,T]$,
    \begin{equation*}
        s_\alpha^\star(t,\cdot)\in \mathcal C^{\beta(\alpha)}(\mathbb R^d;\mathbb R^d)\eqsp,
        \qquad \sup_{t\in[0,T]} \|s_\alpha^\star(t,\cdot)\|_{\mathcal C^{\beta(\alpha)}} \le C_{\alpha,\beta}\eqsp.
    \end{equation*}
\end{assumption}

\begin{assumption}[Exponential terminal signal-to-noise decay]
    \label{ass:exp_rho_decay}
    Let $\rho_T:=a_T/b_T$, where $a_T$ and $b_T$ are defined in \autoref{eq:DLPM_YT}.
    We assume that there exist constants $C_\rho>0$ and $c>0$, independent of $T$, such that
    \begin{equation*}
        |\rho_T| \le C_\rho e^{-cT}\eqsp.
    \end{equation*}
    Equivalently, the contribution of the initial data $Y_0$ to the terminal variable $Y_T$ decays exponentially fast relative to the accumulated $\alpha$-stable noise.
\end{assumption}

The next theorem establishes a non-asymptotic decomposition of the DLPM sampling error into initialization, approximation, and statistical estimation components, mirroring the structure obtained previously for VE-GD.

\begin{theorem}[Sampling error bound for DLPM]
    \label{thm:error-DLPM}
    Let $k_{0\mid A}^{(\alpha)}$ denote the target conditional law at time $0$, and let $\overleftarrow q_{0\mid A}^{\widehat\theta_n}$ be the learned reverse model based on $n$ samples, where the parameter $\widehat\theta_n$ is chosen in a class of neural networks with $m$ parameters.
    Under \autoref{ass:holder_reverse} and \autoref{ass:exp_rho_decay}, for any $\delta\in(0,1)$, with probability at least $1-\delta$,
    \begin{equation}\label{eq:DLPM_sampling_bound}
        \mathbb E_A \left[ \mathrm{TV} \left( k_{0\mid A}^{(\alpha)},
        \overleftarrow q_{0\mid A}^{\widehat\theta_n} \right) \right] \lesssim 
        \underbrace{Tm^{-\beta(\alpha)/d} + T \sqrt{ \frac{\mathrm{Comp}+\log(1/\delta)}{n}
        }}_{\text{training error (approximation+estimation)}}+\underbrace{e^{-cT}}_{\text{initialization error}} \eqsp.
    \end{equation}
\end{theorem}
The three terms in the bound admit a clear interpretation.
The first term is the \emph{\underline{approximation}} error, governed by the Hölder regularity $\beta(\alpha)$ of the reverse object.
The second term captures the \emph{\underline{estimation}} error, which depends on the sample size $n$ and the complexity $\mathrm{Comp}$ of the model class.
The third term corresponds to the \emph{\underline{initialization}} error, which decays exponentially fast thanks to the vanishing signal-to-noise ratio at terminal time.
The training contribution can be written as
\begin{equation*}
    T m^{-\beta(\alpha)/d} + T\sqrt{\frac{\mathrm{Comp}+\log(1/\delta)}{n}}\eqsp.
\end{equation*}
Assuming $\mathrm{Comp}\asymp m$, balancing the approximation and estimation terms yields $m \asymp n^{\frac{d}{2\beta(\alpha)+d}}\eqsp$, and therefore the optimized training rate
\begin{equation*}
    T n^{-\frac{\beta(\alpha)}{2\beta(\alpha)+d}}\eqsp.
\end{equation*}
This coincides with the classical nonparametric rate for estimating a $\beta(\alpha)$-smooth object in dimension $d$; see, e.g., \cite{Tsybakov2009}.
As expected, the rate deteriorates with the dimension $d$ and improves with the regularity $\beta(\alpha)$.
In contrast with the VE-GD bound in \autoref{eq:DDPM_sampling_bound}, the initialization error decreases exponentially fast in $T$, rather than polynomially, which is an excellent initialization rate. In our analysis, this rate follows from \autoref{ass:exp_rho_decay}, under which the weight of $Y_0$ in $Y_T$ vanishes exponentially.

\paragraph{Structural lower bounds.}
As for VE-GD, the initialization, approximation, and estimation terms in \autoref{eq:DLPM_sampling_bound} are individually unavoidable on suitable subclasses.
The initialization error is independent of $n$, while classical approximation and minimax arguments yield lower bounds for the local reverse-object errors.
Moreover, the linear dependence on $T$ is sharp: local kernel errors can accumulate across reverse steps without cancellation, yielding a global error of order $T$ times the local error.
Thus, without additional assumptions, the terms in \autoref{thm:error-DLPM} cannot be improved individually, although this does not establish a minimax lower bound for their sum.
Complete arguments, including a proposition establishing the sharpness of the $T$-accumulation, are provided in \autoref{subsec:supp_lower_bound}.

\subsection{Trade-off between the initialization error and the training error}
\label{subsec:trade_off}

\paragraph{What the comparison does and does not say.}
At first sight, DLPM appears more favorable for heavy-tailed data, with an exponentially decaying initialization error $e^{-cT}$, compared to the polynomial rate $T^{-1/2}$ of VE-GD: this is a genuine advantage of DLPM over the VE formulation considered here. The question is whether it survives the cost of training.

\autoref{thm:error-DDPM} and \autoref{thm:error-DLPM} provide a common \emph{initialization--training} decomposition, allowing a comparison of the mechanisms governing both bounds, without establishing an ordering of actual sampling errors.
The analyses retain model-specific regularity assumptions: Sobolev regularity for the data distribution in VE-GD, and Hölder regularity for the learned reverse object in DLPM. They also differ in their time structure: \autoref{thm:error-DDPM} concerns the continuous-time VE dynamics, whereas \autoref{thm:error-DLPM} concerns a discrete-time chain of $T$ Markov kernels.

Together with the structural lower bounds, this yields the following \emph{mechanistic} conclusion: \emph{DLPM buys faster initialization at the price of a training cost that grows linearly with $T$.}
Since increasing $T$ enables the initialization gain, the training cost may dominate in finite-sample regimes.
Further discussion of the assumptions and limitations of this comparison is provided in \autoref{subsec:supp_comparison_discussion}.

\paragraph{Each bound identifies its model's dominant bottleneck.}
For VE-GD, the dominant bottleneck is the training error, through the bias-variance trade-off near $t=0$ governed by the early-stopping parameter $t_0$: the initialization error decays polynomially with $T$ and becomes secondary once $T$ is moderately large, while the training error is not amplified by $T$.
For DLPM, the initialization error decreases with $T$ whereas the training error increases linearly with it, so that better initialization is paid for in training.
Within the DLPM bound alone, balancing $e^{-cT}$ against $T n^{-\beta(\alpha)/(2\beta(\alpha)+d)}$ yields an optimal horizon $T^\star\asymp c^{-1}\frac{\beta(\alpha)}{2\beta(\alpha)+d}\log n$ (up to lower-order terms), beyond which the training error dominates.
\emph{In finite-sample regimes, this crossover may occur before the initialization advantage becomes significant.}

\paragraph{Two independent sources for the training error gap.}
The difference between the training errors of the two models stems from two orthogonal mechanisms.
\begin{itemize}[leftmargin=*, itemsep=1pt, topsep=1pt]
    \item \emph{$T$-amplification, independent of the regularity of $\pidata$.}
    Even when $\beta(\alpha)=\beta$, the DLPM training error scales linearly with $T$, whereas the VE-GD training error does not.
    By \autoref{prop:lower_bound_T}, this dependence is intrinsic to the reverse chain.
    The VE-GD bound, established in continuous time, does not exhibit this factor; its training error is governed by early stopping at $t_0$.

    \item \emph{Regularity of $\pidata$, independent of $T$.}
    As discussed above, the two models learn different objects, endowed with different regularities: the VE-GD rate depends on the Sobolev regularity $\beta$ of $\pidata$, while the DLPM rate depends on the Hölder regularity $\beta(\alpha)$ of the reverse object $s_\alpha^\star$, with no assumed relationship between them.
    If $\beta(\alpha)<\beta$, DLPM incurs an additional regularity penalty.
\end{itemize}
Our conclusion does not require $\beta(\alpha)<\beta$: even when the regularities coincide, the linear factor $T$ remains and is unavoidable.

For the two formulations considered, our results suggest a trade-off: DLPM can improve initialization while making the reverse dynamics more challenging to learn.

The theory provides \emph{intuition} for why DLPM may be harder to train in finite-sample regimes, which we put to the test through the benchmarks in \autoref{sec:main_bench}.
\section{Benchmark}
\label{sec:main_bench}

We evaluate established heavy-tailed (HT) and light-tailed (LT) diffusion models on synthetic heavy-tailed and real-world datasets to empirically assess our theoretical findings.
All experiments are implemented in PyTorch and conducted on NVIDIA V100 GPUs.
The code is publicly available \href{https://github.com/Diffusion-Research-Lab/0_neurips_2026_heavy_tail_dynamics}{here}.
We compare DDPM-V~\citep{ho2020denoising}, DDIM-V~\citep{song2021denoising}, DLPM~\citep{shariatian2025heavy}, t-EDM~\citep{pandey2025heavytailed}, and Linear Gaussian flow~\citep{lipman2023flowmatching,liu2023flowstraight} on global fit, tail coverage, and rare-class recovery.
DDIM-V and DDPM-V share the same model weights, follow the variance-preserving (VP) formulation, and employ V-prediction.
DLPM and t-EDM zero-network controls retain the sampling dynamics with zeroed network parameters.
Model details and references are given in \autoref{subsec:supp_models_detail}.

\paragraph{Data and training.}
\autoref{tab:data_summary} covers two $150$-dimensional stable targets, wet HRRR precipitation, positive wildfire sizes, and class-imbalanced CIFAR images.
HRRR preprocessing removes exactly all-zero vectors, retaining tiny nonzero observations.
CIFAR tests class-frequency imbalance rather than numerical heavy tails.
Sources and exact splits appear in \autoref{subsec:supp_datasets_detail}.
\begin{table}[htbp]
    \centering
    \small
    \begin{minipage}[c]{0.50\linewidth}
        \centering
        \setlength{\tabcolsep}{5pt}
        \resizebox{\linewidth}{!}{%
        \begin{tabular}{lrrrrr}
            \toprule
            Data & $d$ & Train & Val. & Eval. & $p_{\min}$ \\
            \midrule
            Stable iso. & 150 & 25k & 1k & 5M & $10^{-5}$ \\
            Stable mix. & 150 & 25k & 1k & 5M & $10^{-5}$ \\
            HRRR-E & 150 & 20k & 1k & 79.5k & $10^{-3}$ \\
            HRRR-W & 150 & 20k & 1k & 71.7k & $10^{-3}$ \\
            Wildfire & 1 & 25k & 1k & 1M & $10^{-4}$ \\
            CIFAR100-LT & 3,072 & 15.7k & 1.96k & 1.96k & - \\
            \bottomrule
        \end{tabular}%
        }
    \end{minipage}\hfill
    \begin{minipage}[c]{0.45\linewidth}
        \caption{Dataset budgets (rounded). Eval. counts reference and generated samples per method; MMD$^2$ uses smaller subsets. The maximum quantile is $1-p_{\min}$. Exact counts and preprocessing: \autoref{sec:supp_exp_details}.}
        \label{tab:data_summary}
    \end{minipage}
\end{table}

Vector models share a width-$240$, depth-$4$ MLP; CIFAR models share a two-layer transformer.
Pilots use $16$ epochs and validation selection, with three trials for vector data and one for CIFAR.
Final training uses one seed, AdamW with cosine decay, and up to $720$ epochs with early stopping. Detailed budgets are in \autoref{sec:supp_exp_details}.

\paragraph{Evaluation.}
We report coordinate tail-coverage error (TCE), squared RBF-kernel MMD, and CIFAR rare-label recovery/label TV.
Synthetic TCE uses coordinate magnitudes; real-vector TCE uses raw coordinates, consistently for reference and generated samples.
Test-test is a finite-sample baseline. Definitions and evaluator checks are in \autoref{sec:supp_metrics}.

\begin{figure}[t]
    \centering
    \begin{subfigure}[t]{0.39\linewidth}
        \centering
        \includegraphics[width=\linewidth,trim=0 84bp 0 0,clip]{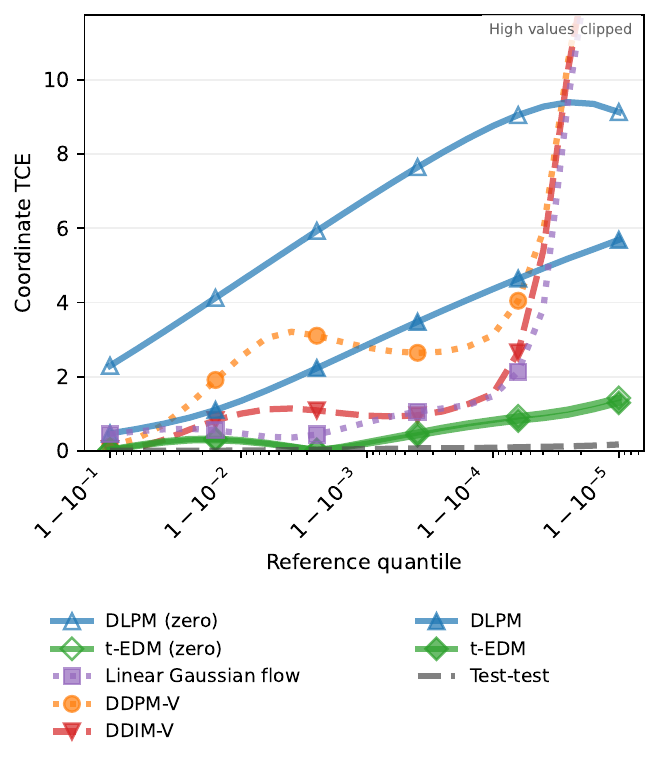}
        \caption{Isotropic $\alpha$-stable.}
    \end{subfigure}\hfill
    \begin{subfigure}[t]{0.39\linewidth}
        \centering
        \includegraphics[width=\linewidth,trim=0 84bp 0 0,clip]{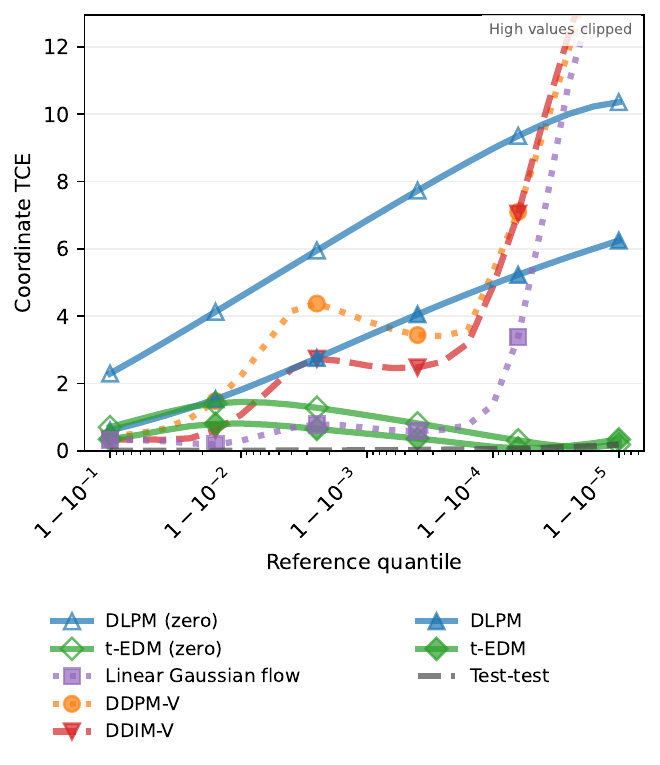}
        \caption{Unbalanced $\alpha$-stable mixture.}
    \end{subfigure}

    \smallskip
    \begin{subfigure}[t]{0.39\linewidth}
        \centering
        \includegraphics[width=\linewidth,trim=0 84bp 0 0,clip]{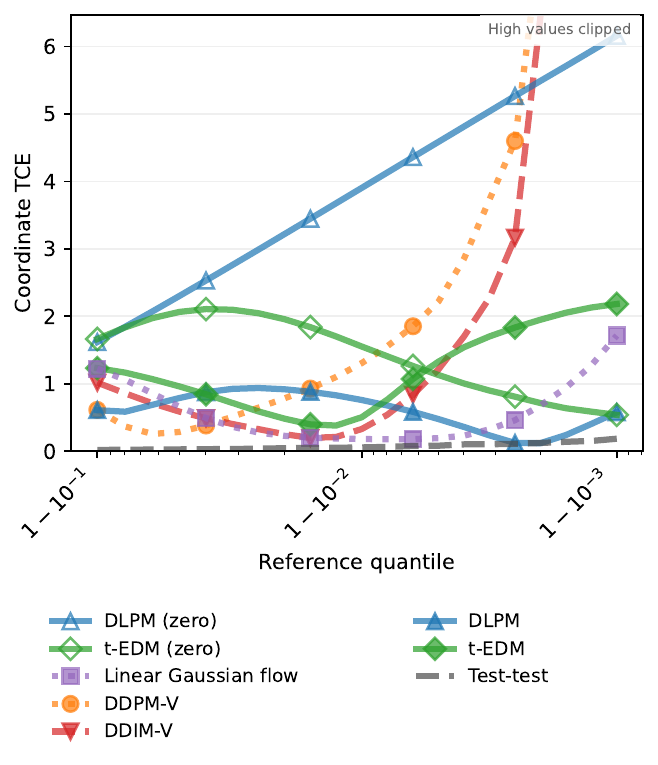}
        \caption{HRRR East, wet hours.}
    \end{subfigure}\hfill
    \begin{subfigure}[t]{0.39\linewidth}
        \centering
        \includegraphics[width=\linewidth,trim=0 84bp 0 0,clip]{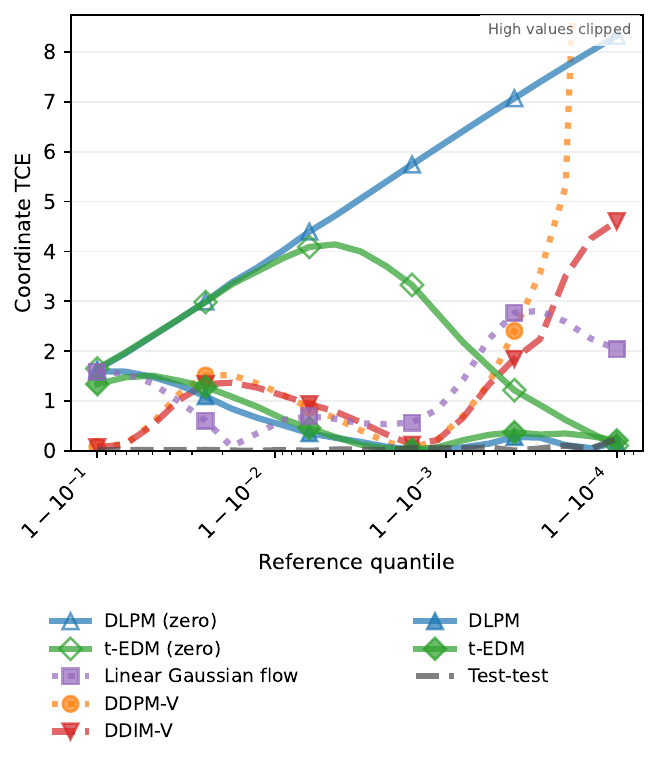}
        \caption{Wildfire size.}
    \end{subfigure}

    \includegraphics[width=0.48\linewidth,trim=15bp 6bp 35bp 289bp,clip]{figures/synthetic_heavy_tail__coordinate_tce__curve.pdf}
    \caption{Coordinate TCE versus reference quantile. Open markers denote zero networks; gray is test-test. High values are clipped where marked. Tail ranges follow \autoref{tab:data_summary}.}
    \label{fig:main_coordinate_tce}
\end{figure}

\subsection{Synthetic and Tabular Results}
\label{subsec:main_synth_heavy_bench}
\label{subsec:main_real_heavy_bench}

In \autoref{fig:main_coordinate_tce}, we first observe that t-EDM barely departs from its zero-network baseline on the synthetic datasets (a) and (b).
This suggests that its performance is largely driven by the inductive bias of its sampling dynamics, rather than by learning the data distribution.
Moreover, when training does affect tail calibration, it does not necessarily improve it, as illustrated in (c).
DLPM exhibits a similar pattern, although it consistently departs from its zero-network baseline.
Despite learning from the data, DLPM shows poor tail calibration on the synthetic datasets (a) and (b), while achieving substantially better tail recovery on HRRR East (c).
The wildfire experiment (d) is particularly informative: in one dimension, with the same 25k training samples, both DLPM and t-EDM clearly improve upon their zero-network controls and recover meaningful tail behavior.
This suggests that their difficulties in higher dimensions may be driven, at least in part, by finite-sample estimation rather than model design alone, see~\autoref{subsec:supp_learning_curves}.
These observations also highlight a limitation of TCE: it measures the recovery of extreme probability mass without necessarily capturing whether the underlying distributional structure has been learned.
A heavy-tailed prior may therefore achieve favorable TCE values without yielding a faithful generative model.

Turning to LT approaches, we observe competitive tail recovery across datasets, at least up to the $1-10^{-2}$ quantile, and even better performance at more extreme quantiles, such as $1-10^{-4}$, in the synthetic cases (a) and (b).
This illustrates that better tail alignment in HT models does not necessarily translate into improved generation.
Consistent with \autoref{thm:error-DLPM} and \autoref{thm:error-DDPM}, learning errors can play a decisive role in sampling performance, particularly in sparsely observed regions such as the tails.

Finally, examining global distributional fidelity through \autoref{fig:main_mmd}, we observe that zero-network baselines fail to capture the underlying data structure.
This further emphasizes that favorable tail calibration alone does not guarantee successful learning of the target distribution.
Indeed, neither TCE nor MMD$^2$ fully captures the extent to which the generative model has learned the relevant structure.
We observe comparable, if not better, recovery of the bulk structure by the LT models.
Notably, linear Gaussian flow matching stands out for its strong performance relative to both HT and other LT alternatives.

\subsection{Rare-Class Recovery on CIFAR100-LT}
\label{subsec:main_cifar}

Heavy-tailed noise has also been proposed to improve robustness to class imbalance and mitigate mode collapse in generative modeling~\citep{shariatian2025heavy}.
We therefore complement the numerical heavy-tail benchmarks with an image-generation experiment assessing minority-class recovery.
We construct CIFAR100-LT by subsampling the CIFAR-100 training images with exponentially decreasing class sizes and an imbalance factor of $10$: the largest class contains ten times as many images as the smallest.
Detailed preprocessing and evaluation settings appear in \autoref{sec:supp_exp_details}.

\begin{figure}[t]
    \centering
    \begin{minipage}[t]{0.5\linewidth}
        \centering
        \vspace{0pt}
        \includegraphics[width=0.75\linewidth]{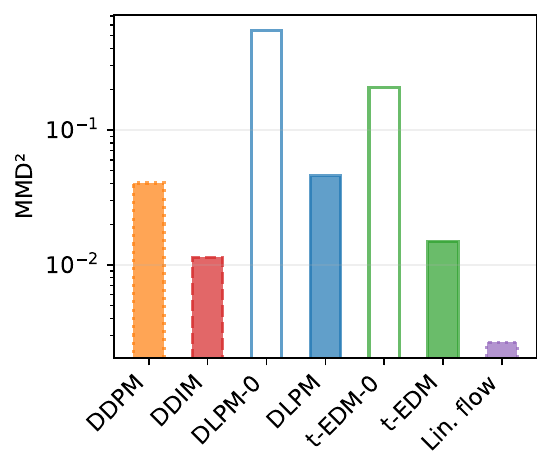}
        \caption{MMD$^2$ on HRRR East, on a log scale; lower is better. "-0" denotes a zero-network control.}
        \label{fig:main_mmd}
    \end{minipage}\hfill
    \begin{minipage}[t]{0.45\linewidth}
        \centering
        \vspace{0pt}
        \captionsetup{type=table}
          \resizebox{0.9\linewidth}{!}{%
  \begin{tabular}{lrr}
    \toprule
    Model & Rare $\uparrow$ (\%) & TV $\downarrow$ (pp) \\
    \midrule
    Reference target & 100.0 & 0.0 \\
    \midrule
    DDPM-V & \textbf{56.3} & 59.4 \\
    DDIM-V & 46.5 & \textbf{56.2} \\
    DLPM & 11.3 & 82.4 \\
    t-EDM & 5.6 & 80.1 \\
    Linear Gaussian flow & 21.1 & 64.6 \\
    \bottomrule
  \end{tabular}%
  }
  \caption{Rare recovery uses validated, quality-filtered rare labels relative to accepted real-test rare mass. TV compares predicted class frequencies.}
  \label{tab:cifar-label-recovery}

    \end{minipage}
\end{figure}

In \autoref{tab:cifar-label-recovery}, DDPM-V achieves the highest rare-label recovery, while DDIM-V has the lowest label TV.
DLPM and t-EDM obtain substantially lower quality-filtered rare-label recovery and larger label TV, indicating poorer recovery of minority-class mass under this evaluation protocol.
Class-frequency recovery alone does not establish perceptual image quality or diagnose mode collapse.
Nevertheless, this experiment highlights that minority-class recovery and numerical tail calibration capture distinct aspects of generative coverage.

Overall, our empirical results suggest that both DLPM and t-EDM struggle to learn the underlying data distribution, particularly in high-dimensional, finite-sample regimes, despite relying on HT noise that better matches the target tails.
Neither model consistently outperforms LT alternatives, consistent with our theoretical insight that improved tail alignment does not necessarily translate into better generation when statistical learning remains the main bottleneck.
Additional results, including sample-size comparisons and a comparison with the released DLPM implementation, are presented in \autoref{sec:supp_exp_addi}.
\section{Conclusion}
\label{sec:ccl}
In this paper, we revisited heavy-tailed (HT) noising in diffusion-based generative modeling, examining whether matching the tail behavior of the noising process to that of the target distribution improves generation.
Our theoretical analysis decomposes the sampling error into initialization and training contributions, further distinguishing approximation and finite-sample estimation effects.
This reveals a \emph{subtle trade-off}: compared with Gaussian VE diffusion, DLPM can reduce initialization error at an exponential rate, but may increase the difficulty of learning the reverse dynamics.
In particular, approximation and estimation errors can accumulate along the reverse process, potentially offsetting the initialization gain.
Our empirical benchmark supports this interpretation: across synthetic and real-world HT datasets, and using metrics assessing both distributional fidelity and tail calibration, HT noising does not consistently improve generative performance.
In several regimes, light-tailed (LT) alternatives remain competitive or outperform DLPM.

Together, these findings suggest that \textbf{better tail alignment does not necessarily translate into better generation, particularly when training data are limited}.
Indeed, a \textbf{central challenge in modeling extreme events is the scarcity of representative tail observations}.
This motivates two complementary research directions, depending on whether additional data can be collected.

\emph{Improving training when additional data are available}: Even with more observations, HT distributions pose optimization challenges due to potentially infinite moments.
Promising directions include robust losses and estimators, gradient clipping, and data transformations for HT models, as well as tail-aware regularization and adaptive noise schedules for LT models.
However, improved training alone cannot compensate for the absence of representative tail observations.

\emph{Correcting generation when additional data are unavailable}: When tail observations are scarce, accurate generation in extreme regions requires additional prior information or structural assumptions.
A promising direction is inference-time correction of pretrained models through initialization, sampling schemes, or \emph{corrective guidance of the reverse dynamics}.
Our analysis suggests that LT models offer a suitable starting point, combining favorable learning properties with guidance toward underrepresented regions.
Although its effectiveness remains unproven, this strategy could improve tail coverage while preserving bulk fidelity without retraining the model.

\clearpage
\newpage

\begin{ack}
We warmly thank Sylvain Le Corff and Gabriel Victorino Cardoso for fruitful discussions, insightful comments, and valuable suggestions that helped improve this work.

The work of Hamza Cherkaoui and H\'el\`ene Halconruy was supported by the French National Research Agency (ANR) under grant ANR-24-CE40-3341 (project DECATTLON).
The work of Antonio Ocello was supported by Hi! PARIS and received government funding managed by the Agence Nationale de la Recherche under the France 2030 program, references (ANR-23-IACL-0005) and it was partially supported by the Labex Ecodec (reference project ANR-11-LABEX-0047).
This work was granted access to the HPC resources of IDRIS under the allocation 2025-AD011016818, provided by GENCI.

\end{ack}

\begingroup
\raggedright
\bibliography{bibliography}
\endgroup

\clearpage
\newpage
\appendix

\section*{Table of contents}
\label{sec:supp_contents}

\begin{tabularx}{\linewidth}{@{}l>{\raggedright\arraybackslash}Xr@{}}
    \autoref{sec:supp_notations} & \nameref*{sec:supp_notations} & \pageref{sec:supp_notations} \\[0.8em]
    \autoref{sec:supp-related-work} & \nameref*{sec:supp-related-work} & \pageref{sec:supp-related-work} \\[0.8em]
    \autoref{sec:supp_proofs} & \nameref*{sec:supp_proofs} & \pageref{sec:supp_proofs} \\[0.8em]
    \autoref{sec:supp_technical} & \nameref*{sec:supp_technical} & \pageref{sec:supp_technical} \\[0.8em]
    \autoref{sec:supp_metrics} & \nameref*{sec:supp_metrics} & \pageref{sec:supp_metrics} \\[0.8em]
    \autoref{sec:supp_exp_details} & \nameref*{sec:supp_exp_details} & \pageref{sec:supp_exp_details} \\[0.8em]
    \autoref{sec:supp_exp_addi} & \nameref*{sec:supp_exp_addi} & \pageref{sec:supp_exp_addi} \\[0.8em]
    \autoref{sec:supp_additional_remarks} & \nameref*{sec:supp_additional_remarks} & \pageref{sec:supp_additional_remarks}
\end{tabularx}\par

\section{Notations}
\label{sec:supp_notations}

\begin{longtable}{@{}p{0.23\linewidth}p{\dimexpr0.77\linewidth-2\tabcolsep\relax}@{}}
    \toprule
    \textbf{Symbol} & \textbf{Description} \\
    \midrule
    \endfirsthead

    \toprule
    \textbf{Symbol} & \textbf{Description} \\
    \midrule
    \endhead

    \bottomrule
    \endfoot

    $\mathbf{1}(\cdot)$ & Indicator function. \\
    $I_d$ & Identity matrix of size $d\times d$. \\
    $a \lesssim b$ & There exists a universal constant $c>0$ \st $a \le c b$. \\
    $h \asymp g$ & There exist universal constants $c_1,c_2>0$ \st $c_1 g \le h \le c_2 g$. \\
    $f = O(g)$ & There exists a constant $c>0$ \st $f \le c g$ for sufficiently large problem size. \\
    $f = \widetilde{O}(g)$ & Same as $O(g)$ up to polylogarithmic factors. \\
    $\mathrm{polylog}(n)$ & polylogarithmic factor in $n$, $\mathrm{polylog}(n)=(\log n)^k,\,k>0$. \\
    $\mathbb{R}^d$ & Euclidean space of dimension $d$. \\
    $\|\cdot\|$ & Euclidean norm. \\
    $\nabla \log p$ & Score function of a density $p$. \\
    $p * q$ & Convolution of two densities. \\
    $\mathcal{L}(X)$ & Law of the random variable $X$. \\
    $\mathbb{E}[\cdot]$, $\mathbb{P}(\cdot)$ & Expectation and probability. \\
    $\mathrm{TV}(\mu,\nu)$ & Total variation distance between probability measures $\mu$ and $\nu$. \\
    $\mathrm{KL}(\mu\,\|\,\nu)$ & Kullback--Leibler divergence between $\mu$ and $\nu$. \\
    $W^{\beta,2}(\mathbb{R}^d)$ & Sobolev space of order $\beta$ with $L^2$ integrability. \\
    $\mathcal{C}^{\beta}(\mathbb{R}^d;\mathbb{R}^d)$ & Hölder space of regularity $\beta$ for vector-valued functions. \\
    $\widehat{\cdot}$ & Empirical or estimated quantity. \\
    $\pidata$, $\mu_\star$ & Target data distribution. \\
    $\gamma>1$ & Tail index of the data distribution. \\
    $\alpha\in(0,2]$ & Tail index of the forward $\alpha$-stable Lévy noise. \\
    $S_\gamma^{\mathbf{i}}(\mu,\sigma I_d)$ & Isotropic $\gamma$-stable distribution with location $\mu$ and scale $\sigma$. \\
    $\beta$, $L$ & Regularity exponent and associated constant. \\
    $T$ & Diffusion horizon. \\
 \\
    $t_0\in(0,T)$ & Early-stopping time used in the DDPM analysis. \\
    $X_t = X_0 + \sqrt{t}\,Z$ & Gaussian forward noising process, with $Z\sim\mathcal N(0,I_d)$. \\
    $\phi_t$ & Density of $\mathcal{N}(0,tI_d)$. \\
    $p_t = \pidata * \phi_t$ & Marginal law of $X_t$ at time $t$. \\
    $\overleftarrow{X}_t$ & Reverse-time diffusion process. \\
    $\widetilde{B}_t$ & Brownian motion driving the reverse SDE. \\
    $s_\theta(x,t)$ & Score network approximating $\nabla\log p_t(x)$. \\
    $\widetilde{Y}$ & Reverse process initialized at the true terminal law. \\
    $\widehat{Y}$ & Practical reverse process initialized from the prescribed prior and driven by $s_\theta$. \\

    $A=(A_1,\dots,A_T)$ & Sequence of $\alpha$-stable Lévy increments. \\
    $S_\alpha$ & Isotropic $\alpha$-stable random variable. \\
    $a_T,b_T$ & Schedule coefficients in $Y_T=a_TY_0+b_TS_\alpha$. \\
    $\rho_T=a_T/b_T$ & Terminal signal-to-noise ratio. \\
    $k_{0:T\mid a}^{(\alpha)}$, $K_a$ & Conditional and marginal forward path laws. \\
    $\overleftarrow{q}_{0:T\mid a}^{\theta}$, $Q_a^\theta$ & Conditional and marginal learned reverse path laws. \\
    $\overleftarrow{q}_T^{\theta}$ & Terminal prior of the learned reverse chain. \\
    $s_\alpha^\star(t,\cdot)$ & True reverse object targeted by the network. \\
    $Y_0,\dots,Y_T$ & State variables of the DLPM forward chain. \\
    $Z=(A,Y_{0:T})$ & Generic training variable. \\

    $\Theta$, $\Theta_m$ & Parameter space and ReLU-network class with $m$ parameters. \\
    $\theta_{\alpha,r}^\star$ & Population minimizer of $\mathcal{L}_\alpha^{\mathrm{TV}}$. \\
    $\widehat{\theta}_n$ & Empirical minimizer of $\widehat{\mathcal{L}}_{n,\alpha}^{\mathrm{TV}}$. \\
    $\ell_\theta^{\mathrm{TV}}(Z)$ & Pointwise TV loss summed over the reverse chain. \\
    $\mathcal{L}_{t-1}^{\mathrm{TV},\alpha}(\theta,a)$ & Local TV loss at step $t$ conditionally on $A=a$. \\
    $\mathcal{L}_\alpha^{\mathrm{TV}}(\theta)$ & Population TV risk. \\
    $\widehat{\mathcal{L}}_{n,\alpha}^{\mathrm{TV}}(\theta)$ & Empirical TV risk over $n$ samples. \\
    $\mathcal{F}^{\mathrm{TV}}$, $\mathcal{F}_t^{\mathrm{TV}}$ & Global and local TV-loss classes. \\
    $\mathfrak{R}_n(\mathcal{F})$ & Empirical Rademacher complexity of $\mathcal{F}$. \\
    $\varepsilon_i$ & I.i.d.\ Rademacher random variables. \\
    $\mathrm{Comp}$ & Generic complexity parameter of the model class. \\
    $\delta\in(0,1)$ & Failure probability. \\

    $\mathcal{E}_{\mathrm{init}}(T)$, $\mathcal{E}_{\mathrm{init}}^{\mathrm{TV}}(T,\alpha)$ & Initialization error. \\
    $\mathcal{E}_{\mathrm{app}}(t_0)$, $\mathcal{E}_{\mathrm{app}}^{\mathrm{TV}}(\alpha)$ & Approximation error. \\
    $\mathcal{E}_{\mathrm{est}}(n,t_0)$, $\mathcal{E}_{\mathrm{est}}^{\mathrm{TV}}(n,\alpha)$ & Estimation error. \\
\end{longtable}

\section{Related work}
\label{sec:supp-related-work}

\paragraph{Heavy-tail generative models.}
Several recent works propose to replace the Gaussian ingredients underlying diffusion and flow-based generative methods with heavier-tailed alternatives.
In particular, \citet{pandey2025heavytailed} introduce $t$-EDM and $t$-Flow, extending EDM \citep{karras2022elucidating} and flow-matching methods \citep{lipman2023flowmatching,liu2023flowstraight}, by replacing Gaussian priors with multivariate Student-$t$ distributions.
\citet{guan2025mirror} combine mirror flow matching with Student-$t$ priors to stabilize constrained generation on convex domains.
More generally, \citet{baule2026generative} extend score-based diffusion to jump-diffusion forward processes driven by Gaussian noise with additional Poisson jumps.
Together, these works follow the same principle: improving tail modeling by replacing Gaussian mechanisms with heavier-tailed counterparts.
Our work revisits this paradigm from a statistical estimation perspective.

In a complementary direction, Denoising L\'{e}vy Probabilistic Models (DLPM) extend the Denoising Diffusion Probabilistic Models (DDPM) \citep{ho2020denoising} construction by replacing Gaussian increments in the forward Markov chain with isotropic $\alpha$-stable noise at each step~\citep{shariatian2025heavy}.
By exploiting the stability of $\alpha$-stable distributions under addition, the forward chain admits an $\alpha$-stable stationary distribution.
The authors then derive a reverse-time Markov chain using detailed balance and adapt the DDPM training machinery through a conditioning argument and a reparameterization trick, leading to a modified denoising objective conditioned on the HT increment injected at each time step.

More recently, several works have explored an alternative paradigm that preserves the standard Gaussian noising process while adapting other components of the generative pipeline to HT settings.
In particular, \citet{yu2026diffusionheavytailedtargets} study the statistical limits of Gaussian score-based diffusion models for HT target distributions, deriving score-estimation and sampling rates that explicitly depend on the tail behavior of the data.
In a different direction, \citet{fassina2026initializationaware} analyze the role of the backward-process initialization and propose a learned initialization strategy that directly minimizes the initialization error without modifying the diffusion dynamics itself.
Together, these works suggest that improving HT generation may not necessarily require changing the noising process, but rather adapting the estimation and initialization procedures underlying diffusion sampling.

\section{\texorpdfstring{Proofs of \autoref{thm:error-DDPM}, \autoref{thm:error-DLPM} and \autoref{prop:lower_bound_T}}{Proofs of the main theorems and initialization lower bound}}
\label{sec:supp_proofs}

\paragraph{Theoretical guarantees.}
Theoretical guarantees for score-based generative models have developed along several complementary directions.
A first line of work studies convergence in strong distributional metrics such as $\alpha$-divergences, Total Variation (TV), and Kullback--Leibler (KL) divergence, typically by controlling the error accumulated along the reverse-time dynamics \citep[see, \eg][]{de2021diffusion,lee2022convergence, chen2023sampling,silveri2024theoretical,conforti2023klconvergence,stephanovitch2026lipschitz}.
Early results established TV guarantees under smoothness assumptions on the score, while more recent analyses obtained KL bounds under milder assumptions on the data distribution; through Pinsker's inequality, such bounds also imply convergence in TV.

A second line of work focuses on Wasserstein distances, which are particularly natural for sampling and estimation problems.
Existing results provide $\wasserstein{1}$ or $\wasserstein{2}$ guarantees under assumptions such as manifold structure, bounded support, score regularity, or log-concavity, with recent refinements further clarifying the role of Lipschitz regularity and generalization in Wasserstein convergence~\citep{gao2025wasserstein,strasman2025analysis,strasman2025wasserstein,silveri2025beyond,stephanovitch2025generalization}.

Our goal is to focus on the statistical \emph{difficulty of the score estimation task} induced by the chosen noising process.
For this reason, we work directly in TV distance, rather than deriving TV guarantees indirectly from KL divergence through Pinsker's inequality.
This choice is particularly natural in the HT setting considered here: when the reference or noising distribution is heavy-tailed, KL-based comparisons with Gaussian counterparts may become infinite or fail to capture the relevant discrepancy.
On the other hand, TV bounds remain well-defined and directly interpretable at the distributional level.
Our analysis is therefore complementary to \citet{fassina2026initializationaware} as both the metric and the resulting error decomposition differ.

\subsection{Gaussian VE diffusion (VE-GD)}

Let now $\pidata$ satisfy \autoref{ass:tail-behavior}.
The corresponding light-tailed diffusion framework based on Gaussian noising processes is presented in \autoref{subsec:theory_lighttail}.

\underline{\textbf{Error decomposition}.}
Because $p_T = \pidata * \phi_T$ remains heavy-tailed, the KL divergence with a Gaussian prior may be infinite (this is the case when $\gamma\le2$, since $p_T$ then has no finite second moment).

It is therefore natural to measure the error in total variation.
For an early stopping time $t_0\in(0,T)$, define $p_{t_0}=\pidata * \phi_{t_0}$.
Let $\widetilde Y$ denote the reverse process initialized from $p_T$, and $\widehat Y$ the practical reverse process initialized from $\phi_T=\mathcal N(0,T I_d)$ and driven by $s_\theta$.
Then
\begin{equation*}
    \mathrm{TV} \big( \pidata, \mathcal L(\widehat Y_{T-t_0}) \big)
    \le \underbrace{ \mathrm{TV}(\pidata,p_{t_0})}_{\mathcal E_{\mathrm{app}}(t_0)}
    + \underbrace{ \mathrm{TV} \big( p_{t_0}, \mathcal L(\widetilde Y_{T-t_0}) \big) }_{\mathcal E_{\mathrm{est}}(n,t_0)}
    + \underbrace{ \mathrm{TV} \big( p_T,\phi_T \big)}_{\mathcal E_{\mathrm{init}}(T)}\eqsp.
\end{equation*}

\medskip

\underline{\textbf{Approximation error} $\mathcal E_{\mathrm{app}}(t_0)$.}
The term $\mathcal E_{\mathrm{app}}(t_0)$ quantifies the bias introduced by Gaussian smoothing.
It reflects the discrepancy between the original density and its regularized version $p_{t_0}$.
Under Sobolev regularity and polynomial tails, \cite{yu2026diffusionheavytailedtargets} show that
\begin{equation*}
    \mathcal E_{\mathrm{app}}(t_0) \lesssim t_0^{ \frac{\beta(\gamma+1)} {d+2(\gamma+1)+2\beta}}\eqsp.
\end{equation*}
This rate arises from a balance between local smoothing (controlled by $\beta$) and tail mass (controlled by $\gamma$).

\medskip

\underline{\textbf{Estimation error} $\mathcal E_{\mathrm{est}}(n,t_0)$.}
The term $\mathcal E_{\mathrm{est}}(n,t_0)$ captures the error due to score estimation.
It measures how well the learned score $s_\theta$ approximates $\nabla \log p_t$ along the reverse trajectory.
It can be bounded in terms of the cumulative $\mathbb L^2(p_t)$-error:
\begin{equation*}
    \mathcal E_{\mathrm{est}}(n,t_0) \lesssim \left( \int_{t_0}^T \mathbb E \left[ \int \| s_\theta(x,t)-\nabla\log p_t(x) \|^2 p_t(x)\,\rmd x \right] \rmd t \right)^{1/2}\eqsp.
\end{equation*}
For heavy-tailed targets, this leads to the nonparametric rate (see \cite{yu2026diffusionheavytailedtargets}, Theorem 2)
\begin{equation*}
    \mathcal E_{\mathrm{est}}(n,t_0) \lesssim \operatorname{polylog}(n)\, n^{-\frac{\gamma+1}{2(d+\gamma+1)}}\, t_0^{-\frac{d(\gamma+1)}{4(d+\gamma+1)}}\eqsp.
\end{equation*}

\medskip

\underline{\textbf{Initialization error} $\mathcal E_{\mathrm{init}}(T)$.}
The initialization error $\mathcal E_{\mathrm{init}}(T)$ reflects the mismatch between the true terminal distribution $p_T$ and the Gaussian prior used in practice.
Since $\gamma>1$, the first moment is finite, and one obtains
\begin{equation*}
    \mathcal E_{\mathrm{init}}(T) \lesssim T^{-1/2}\eqsp.
\end{equation*}
This term decreases with the diffusion horizon and is independent of the learning procedure.

\medskip

\underline{\textbf{Final bound}.}
Combining the three contributions, we obtain
\begin{equation*}
    \mathrm{TV} \big( \pidata, \mathcal L(\widehat Y_{T-t_0}) \big)
    \lesssim t_0^{ \frac{\beta(\gamma+1)} {d+2(\gamma+1)+2\beta} }
    + \operatorname{polylog}(n)\, n^{-\frac{\gamma+1}{2(d+\gamma+1)}} \, t_0^{-\frac{d(\gamma+1)}{4(d+\gamma+1)}} + T^{-1/2}\eqsp.
\end{equation*}

\subsection{DLPM}
\label{subsec:supp_DLPM_proof}

The corresponding heavy-tailed generative setting based on $\alpha$-stable noising processes is introduced in \autoref{subsec:theory_heavytail}.

\underline{\textbf{Decomposition of the error}.}
Since $k_{0| a}^{(\alpha)}$ and $\overleftarrow q_{0| a}^{\theta}$ are the $Y_0$-marginals of $K_a$ and $Q_a^\theta$, respectively, the data-processing inequality for total variation yields
\begin{equation}\label{eq:DLPM_TV_decomposition_1}
    \mathrm{TV} \big( k_{0| a}^{(\alpha)}, \overleftarrow q_{0| a}^{\theta} \big) \le \mathrm{TV}(K_a,Q_a^\theta).
\end{equation}
Using contraction of total variation under Markov kernels and a telescoping argument for inhomogeneous chains, we obtain
\begin{align}\label{eq:DLPM_TV_decomposition_2}
    \mathrm{TV}(K_a,Q_a^\theta) \nonumber & \le \mathrm{TV} \big( k_{T| a}^{(\alpha)}, \overleftarrow q_T^\theta \big) \\
    \nonumber & \quad + \sum_{t=1}^T \mathbb E_{K_a} \left[ \mathrm{TV} \big( k_{t-1| t,a}^{(\alpha)}(\cdot| Y_t,a), \overleftarrow q_{t-1| t,a}^{\theta}(\cdot| Y_t,a) \big) \right] \\
    & = \mathrm{TV} \big( k_{T| a}^{(\alpha)}, \overleftarrow q_T^\theta \big)+\sum_{t=1}^T\mathcal{L}_{t-1}^{\mathrm{TV},\alpha}(\theta,a)
\end{align}
where for each $t\in\{1,\dots,T\}$ and $a\in\mathbb R^T$, we define the local TV loss
\begin{equation*}
    \mathcal L_{t-1}^{\mathrm{TV},\alpha}(\theta,a) := \mathbb E_{K_a} \left[ \mathrm{TV} \big( k_{t-1| t,a}^{(\alpha)}(\cdot| Y_t,a), \overleftarrow q_{t-1| t,a}^{\theta}(\cdot| Y_t,a) \big) \right]\eqsp,
\end{equation*}
    and the expectation is taken with respect to the true path law $K_a$.
    The population TV risk is then defined as
\begin{equation*}
    \mathcal L_\alpha^{\mathrm{TV}}(\theta) := \sum_{t=1}^T \mathbb E_A \left[ \mathcal L_{t-1}^{\mathrm{TV},\alpha}(\theta,A) \right]\eqsp.
\end{equation*}
Equivalently, introducing a generic data variable $Z:=(A,Y_{0:T})\sim K_A \otimes \mathbb P_A$.
We can write
\begin{equation*}
    \mathcal L_\alpha^{\mathrm{TV}}(\theta) = \mathbb E\big[\ell_\theta^{\mathrm{TV}}(Z)\big],
    \qquad \ell_\theta^{\mathrm{TV}}(Z) := \sum_{t=1}^T \mathrm{TV} \big( k_{t-1| t,A}^{(\alpha)}(\cdot| Y_t,A), \overleftarrow q_{t-1| t,A}^{\theta}(\cdot| Y_t,A) \big)\eqsp.
\end{equation*}
Given i.i.d.\ samples $Z_1,\dots,Z_n$, the empirical TV risk is
\begin{equation*}
    \widehat{\mathcal L}_{n,\alpha}^{\mathrm{TV}}(\theta) := \frac1n \sum_{i=1}^n \ell_\theta^{\mathrm{TV}}(Z_i)\eqsp.
\end{equation*}
Their respective minimizers are given by
\begin{equation*}
    \theta_{\alpha,r}^\star \in \arg\min_{\theta\in\Theta} \mathcal L_{\alpha}^{\mathrm{TV}}(\theta) \quad \text{and} \quad \widehat\theta_n \in \arg\min_{\theta\in\Theta} \widehat{\mathcal L}_{n,\alpha}^{\mathrm{TV}}\eqsp.
\end{equation*}
Then, plugging $\theta=\widehat\theta_n$ into the previous inequalities in \autoref{eq:DLPM_TV_decomposition_1} and \autoref{eq:DLPM_TV_decomposition_2} gives
\begin{align*}
    \mathbb E_A \left[\mathrm{TV}\big( k_{0| A}^{(\alpha)}, \overleftarrow q_{0| A}^{\widehat\theta_n} \big) \right]
    &\le \mathbb E_A \left[\mathrm{TV}(K_A,Q_A^{\widehat\theta_n}) \right] \\
    & \le \underbrace{ \mathbb E_A\big[\mathrm{TV} \big( k_{T|A}^{(\alpha)}, \overleftarrow q_T^{\widehat\theta_n} \big)\Big] }_{\mathcal E_{\mathrm{init}}^{\mathrm{TV}}(T,\alpha)}
    + \underbrace{ \mathcal L_{\alpha}^{\mathrm{TV}}(\theta_{\alpha,r}^\star) }_{\mathcal E_{\mathrm{app}}^{\mathrm{TV}}(\alpha)}
    + \underbrace{ \left[ \mathcal L_{\alpha}^{\mathrm{TV}}(\widehat\theta_n) - \mathcal L_{\alpha}^{\mathrm{TV}}(\theta_{\alpha,r}^\star) \right]}_{\mathcal E_{\mathrm{est}}^{\mathrm{TV}}(n,\alpha)}\eqsp.
\end{align*}

\medskip

\underline{\textbf{Approximation error $\mathcal E_{\mathrm{app}}^{\mathrm{TV}}(\alpha)$}.}
In the DLPM setting, the forward dynamics involve convolution with an $\alpha$-stable law.
For any positive time $t>0$, this induces a smoothing effect: the resulting density belongs to fractional Hölder (or Sobolev) spaces.
Since the reverse object $s_\alpha^\star$ is constructed from this regularized density (e.g.\ via derivatives or score-type expressions), it inherits a positive degree of Hölder regularity.
This motivates the following assumption.

Let $\Theta_m$ be a class of ReLU neural networks with $m$ parameters and sufficient depth.
A standard approximation result for ReLU neural networks states that for any $f\in\mathcal C^\beta([0,1]^d)$, there exists a network $f_m$ with $m$ parameters such that
\begin{equation*}
    \|f-f_m\|_{\mathbb L^\infty([0,1]^d)} \lesssim m^{-\beta/d} \eqsp,
\end{equation*}
up to logarithmic factors (see \cite{schmidt2020nonparametric}). Applying this result componentwise to
$s_\alpha^\star(t,\cdot)$ yields a network $s_\theta$ such that
\begin{equation*}
    \sup_{t\in[0,T]} \|s_\theta(t,\cdot)-s_\alpha^\star(t,\cdot)\|_{\mathbb L^\infty} \lesssim m^{-\beta(\alpha)/d}\eqsp.
\end{equation*}
Therefore, for any distribution of $(Y_t,A)$,
\begin{equation*}
    \mathbb E\big[ \|s_\theta(t,Y_t,A)-s_\alpha^\star(t,Y_t,Y_0,A)\|^2\big]
    \le \|s_\theta-s_\alpha^\star\|_{\mathbb L^\infty}^2
    \lesssim m^{-2\beta(\alpha)/d}\eqsp.
\end{equation*}
Then,
\begin{equation*}
    \inf_{\theta\in\Theta_m} \left( \mathbb E \|s_\theta(t,Y_t,A)-s_\alpha^\star(t,Y_t,Y_0,A)\|^2 \right)^{1/2} \lesssim  m^{-\beta(\alpha)/d}\eqsp,
\end{equation*}
up to logarithmic factors. The implicit constant depends on $d,\beta(\alpha)$ and $C_{\alpha,\beta}$, but not on $m$.

\noindent \underline{\textbf{Estimation error $\mathcal E_{\mathrm{est}}^{\mathrm{TV}}(n,\alpha)$}.}
Define the loss class
\begin{equation*}
    \mathcal F^{\mathrm{TV}} = \{\ell_\theta^{\mathrm{TV}}:\theta\in\Theta\}\eqsp.
\end{equation*}
Since the TV loss is uniformly bounded, the class $\mathcal F^{\mathrm{TV}}$ is bounded as well.
Applying the Rademacher complexity bound of \autoref{th:Rademacher_bound}, we obtain that, with probability at least $1-\delta$,
\begin{equation*}
    \mathcal E_{\mathrm{est}}^{\mathrm{TV}}(n,\alpha) \lesssim \mathfrak R_n(\mathcal F^{\mathrm{TV}}) + T\sqrt{\frac{\log(1/\delta)}{n}}\eqsp.
\end{equation*}
To control the Rademacher complexity, we use
\begin{equation*}
    \ell_\theta^{\mathrm{TV}}(Z) = \sum_{t=1}^T \ell_{t,\theta}^{\mathrm{TV}}(Z),
    \quad \text{with}
    \quad \ell_{t,\theta}^{\mathrm{TV}}(Z) := \mathrm{TV} \left( k_{t-1\mid t,A}^{(\alpha)}(\cdot\mid Y_t,A), \overleftarrow q_{t-1\mid t,A}^{\theta}(\cdot\mid Y_t,A) \right)\eqsp.
\end{equation*}
By subadditivity of Rademacher complexity,
\begin{equation*}
    \mathfrak R_n(\mathcal F^{\mathrm{TV}}) \le \sum_{t=1}^T \mathfrak R_n(\mathcal F_t^{\mathrm{TV}})\eqsp,
\end{equation*}
where $\mathcal F_t^{\mathrm{TV}} := \{\ell_{t,\theta}^{\mathrm{TV}}:\theta\in\Theta\}$.
Assume that each class $\mathcal F_t^{\mathrm{TV}}$ satisfies
\[
    \mathfrak R_n(\mathcal F_t^{\mathrm{TV}}) \lesssim \sqrt{\frac{\mathrm{Comp}}{n}},
\]
for some complexity parameter $\mathrm{Comp}$ depending on the model class (e.g.\ number of parameters, norm constraints, or network depth).

Then
\[
    \mathfrak R_n(\mathcal F^{\mathrm{TV}}) \lesssim T\sqrt{\frac{\mathrm{Comp}}{n}},
\]
and therefore
\begin{equation*}
    \mathcal E_{\mathrm{est}}^{\mathrm{TV}}(n,\alpha) \lesssim T\sqrt{\frac{\mathrm{Comp}}{n}} + T\sqrt{\frac{\log(1/\delta)}{n}}\eqsp.
\end{equation*}

Combining terms yields
\begin{equation*}
    \mathcal E_{\mathrm{est}}^{\mathrm{TV}}(n,\alpha) \lesssim T \sqrt{ \frac{\mathrm{Comp}+\log(1/\delta)}{n} }\eqsp.
\end{equation*}

\medskip

\underline{\textbf{Initialization error} $\mathcal E_{\mathrm{init}}^{\mathrm{TV}}(T,\alpha)$}.
We assume that the terminal prior is chosen as
\begin{equation*}
    \overleftarrow q_T^\theta = \mathcal L(b_T S_\alpha)\eqsp,
\end{equation*}
which matches the dominant noise component of $Y_T$ in \autoref{eq:DLPM_YT}.
Introducing $\rho_T := a_T/b_T$, we may rewrite
\begin{equation*}
    Y_T = b_T(\rho_T Y_0 + S_\alpha)\eqsp.
\end{equation*}
Since total variation is invariant under invertible linear transformations, it follows that
\[
    \mathrm{TV}\big(\mathcal L(Y_T), \mathcal L(b_T S_\alpha)\big) = \mathrm{TV}\big(\mathcal L(\rho_T Y_0 + S_\alpha), \mathcal L(S_\alpha)\big).
\]
Conditioning on $Y_0$ and using convexity of total variation yields
\[
    \mathcal E_{\mathrm{init}}^{\mathrm{TV}}(\alpha) = \mathbb E \left[ \mathrm{TV} \left( p_\alpha(\cdot-\rho_T Y_0), p_\alpha \right) \right],
\]
where $p_\alpha$ denotes the density of $S_\alpha$.

To control this quantity, we use a first-order Taylor expansion.
For any $y\in\mathbb R^d$,
\[
    p_\alpha(x-\rho y)-p_\alpha(x) = - \int_0^1 \rho\, y \cdot \nabla p_\alpha(x-s\rho y)\,\rmd s,
\]
which implies
\[
    \mathrm{TV} \big( p_\alpha(\cdot-\rho y), p_\alpha \big) \le \frac{|\rho|\,\|y\|}{2} \int_{\mathbb R^d} \|\nabla p_\alpha(x)\|\,\rmd x.
\]

The integrability condition
\[
    \int_{\mathbb R^d}\|\nabla p_\alpha(x)\|\,\rmd x<\infty
\]
holds for $\alpha$-stable densities (which are smooth with polynomially decaying derivatives; see \cite{sato1999levy}).
Under \autoref{ass:exp_rho_decay}, the initialization bound
\begin{equation*}
    \mathcal E_{\mathrm{init}}^{\mathrm{TV}}(\alpha) \lesssim |\rho_T|\,\mathbb E\|Y_0\|
\end{equation*}
implies, since $\mathbb E\|Y_0\|<\infty$ ($Y_0$ has $\gamma$-stable tails with $\gamma\in(1,2)$),
\begin{equation*}
    \mathcal E_{\mathrm{init}}^{\mathrm{TV}}(\alpha) = \mathrm O_{\alpha,\gamma}(e^{-cT})\eqsp.
\end{equation*}

\medskip

\noindent \underline{\textbf{Final bound}}.
Combining the initialization, approximation, and estimation bounds, we obtain, with probability at least $1-\delta$,
\begin{equation*}
    \mathbb E_A \left[ \mathrm{TV} \left( k_{0\mid A}^{(\alpha)}, \overleftarrow q_{0\mid A}^{\widehat\theta_n} \right) \right] \lesssim
    e^{-cT} + T\,m^{-\beta(\alpha)/d} + T \sqrt{ \frac{\mathrm{Comp}+\log(1/\delta)}{n}}.
\end{equation*}

\subsection{\texorpdfstring{Proof of \autoref{prop:lower_bound_T}}{Proof of the initialization lower bound}}
\label{subsec:proof_prop_T}
\underline{\textbf{Construction}.}
Set $\Omega=\{0,1,\dots,T\}\cup\{\ast\}$ and $\mu_T=\delta_T$.
For each step $t\in\{1,\dots,T\}$, define the true kernel
\begin{equation*}
    Q_t(x,\cdot) =
    \begin{cases}
        \delta_{x-1} & \text{if } x\in\{1,\dots,T\}\eqsp,\\
        \delta_x & \text{if } x\in\{0,\ast\}\eqsp.
    \end{cases}
\end{equation*}
Each $Q_t$ decrements the state by one, the states $0$ and $\ast$ being absorbing.
Starting from $\mu_T=\delta_T$, the true chain moves deterministically along $T\to T-1\to\cdots\to0$, so that $\mu_TQ_T\cdots Q_1=\delta_0$.
In particular, the true chain is at state $t$ exactly when the kernel of step $t$ is applied.
The learned kernels are obtained by introducing a single \emph{error channel} at each step:
\begin{equation*}
    \widehat Q_t(x,\cdot) =
    \begin{cases}
        (1-\varepsilon)\,\delta_{t-1}+\varepsilon\,\delta_\ast & \text{if } x=t\eqsp,\\
        Q_t(x,\cdot) & \text{if } x\neq t\eqsp.
    \end{cases}
\end{equation*}
The kernels $\widehat Q_t$ and $Q_t$ coincide except at state $x=t$, where $\widehat Q_t$ sends a fraction $\varepsilon$ of the mass to the absorbing state $\ast$.
Hence
\begin{equation*}
    \sup_{x\in\Omega}\mathrm{TV}\big(Q_t(x,\cdot),\widehat Q_t(x,\cdot)\big) = \mathrm{TV}\big(\delta_{t-1},(1-\varepsilon)\delta_{t-1}+\varepsilon\delta_\ast\big) = \varepsilon\eqsp,
\end{equation*}
which proves the first claim.

\medskip

\noindent \underline{\textbf{Propagation of the error}}.
We show by induction that, for every $k\in\{0,\dots,T\}$,
\begin{equation}\label{eq:prop_T_induction}
    \mu_T\widehat Q_T\cdots\widehat Q_{T-k+1} = (1-\varepsilon)^k\,\delta_{T-k} + \big(1-(1-\varepsilon)^k\big)\,\delta_\ast\eqsp,
\end{equation}
with the convention that the empty product is the identity.
For $k=0$, \autoref{eq:prop_T_induction} reduces to $\mu_T=\delta_T$.
Assume that \autoref{eq:prop_T_induction} holds for some $k<T$ and apply $\widehat Q_{T-k}$.
The mass located at $\ast$ remains there since $\ast$ is absorbing, while the mass $(1-\varepsilon)^k$ located at $T-k$ meets the error channel of step $T-k$ and is mapped to $(1-\varepsilon)\delta_{T-k-1}+\varepsilon\delta_\ast$.
Therefore,
\begin{align*}
    \mu_T\widehat Q_T\cdots\widehat Q_{T-k}
    &= (1-\varepsilon)^{k+1}\,\delta_{T-k-1} + \big(1-(1-\varepsilon)^k+\varepsilon(1-\varepsilon)^k\big)\,\delta_\ast\\
    &= (1-\varepsilon)^{k+1}\,\delta_{T-k-1} + \big(1-(1-\varepsilon)^{k+1}\big)\,\delta_\ast\eqsp,
\end{align*}
which completes the induction. Taking $k=T$ gives
\begin{equation*}
    \mu_T\widehat Q_T\cdots\widehat Q_1 = (1-\varepsilon)^T\,\delta_0 + \big(1-(1-\varepsilon)^T\big)\,\delta_\ast\eqsp.
\end{equation*}
Since $\delta_0$ and $\delta_\ast$ are mutually singular,
\begin{equation*}
    \mathrm{TV}\big(\mu_TQ_T\cdots Q_1,\ \mu_T\widehat Q_T\cdots\widehat Q_1\big) = 1-(1-\varepsilon)^T\eqsp.
\end{equation*}
The error channels are disjoint, in the sense that each of them is activated exactly once along the true trajectory, and every discrepancy pushes mass in the same direction, towards $\ast$.
The local errors therefore accumulate without cancellation.

\medskip

\noindent\underline{\textbf{Lower bound}}.
Using $1-\varepsilon\le e^{-\varepsilon}$, we get $1-(1-\varepsilon)^T\ge1-e^{-T\varepsilon}$.
If $T\varepsilon\le1$, the concavity of $x\mapsto1-e^{-x}$ gives $1-e^{-x}\ge(1-e^{-1})x\ge x/2$ on $[0,1]$, hence $1-e^{-T\varepsilon}\ge T\varepsilon/2$.
If $T\varepsilon>1$, then $1-e^{-T\varepsilon}>1-e^{-1}>1/2$.
In both cases,
\begin{equation*}
    1-(1-\varepsilon)^T\ge\tfrac12\min\{1,T\varepsilon\}\eqsp.
\end{equation*}
Conversely, the telescoping argument of \autoref{eq:DLPM_TV_decomposition_2} gives $1-(1-\varepsilon)^T\le T\varepsilon$, so that the linear accumulation is sharp up to a universal constant.
\hfill$\square$

\subsection{Comparison of the VE-GD and DLPM bounds.}
We now compare the two bounds through a common decomposition into \emph{initialization}, \emph{approximation}, and \emph{estimation} errors, in order to highlight the respective strengths and limitations of VE-GD and DLPM.\\
\noindent \textbf{VE-GD bound.}
\begin{equation*}
    \mathrm{TV} \big( \pidata, \mathcal L(\widehat Y_{T-t_0}) \big)
    \lesssim\; \underbrace{T^{-1/2}}_{\text{initialization error}}
    + \underbrace{ t_0^{ \frac{\beta(\gamma+1)} {d+2(\gamma+1)+2\beta}}
    + \operatorname{polylog}(n)\, n^{-\frac{\gamma+1}{2(d+\gamma+1)}} \, t_0^{-\frac{d(\gamma+1)}{4(d+\gamma+1)}}  }_{\text{training error}}\eqsp.
\end{equation*}

\medskip

\noindent \textbf{DLPM bound}.
\begin{equation*}
    \mathbb E_A \left[ \mathrm{TV} \left( k_{0\mid A}^{(\alpha)}, \overleftarrow q_{0\mid A}^{\widehat\theta_n} \right) \right]
    \lesssim \underbrace{e^{-cT}}_{\text{initialization error}}
    + \underbrace{ T\,m^{-\beta(\alpha)/d} + T \sqrt{ \frac{\mathrm{Comp}+\log(1/\delta)}{n}} }_{\text{training error}}\eqsp.
\end{equation*}

\medskip

\noindent We now compare each source of error in turn.

\medskip

\noindent \emph{Initialization}.
The initialization term measures how well the terminal distribution matches the chosen prior.
In VE-GD, this term decays only polynomially as $T^{-1/2}$, because in the VE dynamics \autoref{eq:general-fwd-dynamics} the initial condition is never rescaled: $p_T$ is a mixture of $\mathcal N(x,TI_d)$, $x\sim\pidata$, whose discrepancy with $\phi_T$ only vanishes through the growth of the variance.
In contrast, DLPM achieves an exponential decay $e^{-cT}$, because its noise schedule makes the weight of $Y_0$ in $Y_T$ vanish exponentially relative to the injected noise (\autoref{ass:exp_rho_decay}).
This exponential rate is a substantial advantage of DLPM over the VE-GD LT model.
Our comparison concerns these two representative formulations, Gaussian VE diffusion and stable-noise DLPM.

\medskip

\noindent \emph{Training error (approximation + estimation)}.
The training error can be decomposed into two components: an \emph{approximation error}, which measures how well the model class can
represent the reverse dynamics, and an \emph{estimation error}, which reflects the statistical difficulty of learning from finite data.
We first consider the approximation component.
In DLPM, this error scales as $T m^{-\beta(\alpha)/d}$, reflecting the accumulation of local approximation errors along the $T$ reverse
steps.
In practice, $T$ must be large to ensure that the forward process sufficiently mixes (so that the terminal distribution is close to the prior).
As a consequence, even moderate local approximation errors can accumulate and dominate the bound.

By contrast, the VE-GD bound, established in continuous time, does not exhibit this accumulation.
The approximation error is controlled through the early-stopping parameter $t_0$, which balances bias and variance, and does not scale with $T$.

A similar phenomenon occurs for the estimation component.
In DLPM, the loss is a sum of local contributions along the reverse chain, which leads to an estimation error of the form
$T\sqrt{(\mathrm{Comp}+\log(1/\delta))/n}$ reflecting the accumulation of statistical fluctuations across the $T$ steps.
Since each step involves learning a conditional distribution, the overall estimation error grows linearly with $T$.
By contrast, the VE-GD bound does not exhibit such accumulation.
The estimation error is controlled globally through the early-stopping parameter $t_0$, leading to a
rate of the form $n^{-\frac{\gamma+1}{2(d+\gamma+1)}}$, without any multiplicative dependence on $T$.
As a result, even though this rate may deteriorate with the tail index $\gamma$, it does not suffer from the same amplification mechanism , so that the corresponding bound does not grow with $T$.

\medskip

\noindent \textbf{Summary}.
Overall, while DLPM benefits from a significantly improved initialization, its \emph{training error} (defined as the sum of approximation and estimation errors) scales linearly with $T$.
Since $T$ must be large to ensure good initialization, these errors accumulate along the reverse chain and can dominate the overall bound.
As a result, even though DLPM enjoys a faster initialization rate than Gaussian VE diffusion, VE-GD may be preferable in practice when the training error is the main bottleneck.

\section{Useful technical results}
\label{sec:supp_technical}

We recall a standard uniform deviation bound for bounded function classes.
Let $\mathcal F $ be a class of functions.
Define the empirical Rademacher complexity
\begin{equation}
    \label{eq:Rademacher_def}
    \mathfrak R_n(\mathcal F) = \mathbb E_{\varepsilon,Z} \left[ \sup_{f\in\mathcal F} \frac{1}{n} \sum_{i=1}^n \varepsilon_i f(Z_i) \right],
\end{equation}
where $(\varepsilon_i)_{i=1}^n$ are i.i.d.\ Rademacher variables independent of the sample $(Z_i)_{i=1}^n$.
\begin{theorem}[Uniform deviation bound for bounded classes]
    \label{th:Rademacher_bound} Assume that the class $\mathcal F$ is uniformly bounded, namely there exists $B>0$ such that
    \begin{equation*}
        \sup_{f\in\mathcal F} \|f\|_{\infty} \le B\eqsp.
    \end{equation*}
    Then, with probability at least $1-\delta$,
    \begin{equation*}
        \sup_{f\in\mathcal F} \left| \mathbb E[f(Z)] - \frac{1}{n} \sum_{i=1}^n f(Z_i) \right| \le 2 \,\mathfrak R_n(\mathcal F) + C B \sqrt{\frac{\log(1/\delta)}{n}}\eqsp,
    \end{equation*}
    for some universal constant $C>0$.
\end{theorem}

\section{Metrics}
\label{sec:supp_metrics}

We evaluate vector distributions in their original coordinates after undoing any training standardization.
Lower scores are better unless stated otherwise.
Main vector results are single-final-run estimates, not means or standard deviations over repeated final training runs.

\subsection{Coordinate Tail Coverage}
\label{subsec:supp_tce}

Let $x_1,\ldots,x_n$ be reference observations and $y_1,\ldots,y_m$ generated observations in $\mathbb{R}^d$. Define $g(x)=|x|$ coordinatewise for the synthetic benchmarks and $g(x)=x$ for HRRR and wildfire. For each reference quantile $q$ and coordinate $j$, compute
\begin{align*}
    u_j(q)&=\widehat{Q}_q\bigl(\{g(x_i)_j\}_{i=1}^n\bigr),\\
    r_j(q)&=\frac{1}{n}\sum_{i=1}^n\mathbf{1}\{g(x_i)_j>u_j(q)\},\qquad
    s_j(q)=\frac{1}{m}\sum_{i=1}^m\mathbf{1}\{g(y_i)_j>u_j(q)\}.
\end{align*}
The coordinate-TCE curve is
\begin{equation}
    C(q)=\frac{1}{d}\sum_{j=1}^d\left|\log\frac{s_j(q)+\varepsilon}{r_j(q)+\varepsilon}\right|,
    \qquad \varepsilon=10^{-12}.
    \label{eq:empirical_tce}
\end{equation}
Thus reference and generated samples receive the same transformation.
The denominator is the empirical reference exceedance mass, rather than exactly $1-q$.
This targets tail calibration directly~\citep{allen2025tailcalibration}; an identical reference/generated sample has zero error.

Each curve uses $21$ logarithmically spaced exceedance probabilities $p=1-q$, from $10^{-1}$ to $p_{\min}$.
The minimum is $10^{-5}$ for the synthetic benchmarks, $10^{-3}$ for HRRR, and $10^{-4}$ for wildfire.
The reference quantile axis is displayed on a logit scale.
The test-test baseline compares two disjoint halves of the reference sample using the same metric; it is a finite-sample diagnostic, not a confidence interval.
Displayed curves may clip high values for readability; the scores themselves are not clipped.

\subsection{Squared RBF-Kernel MMD}
\label{subsec:supp_mmd}

We report the biased empirical squared maximum mean discrepancy~\citep{gretton2012mmd}, including diagonal kernel terms:
\begin{equation}
    \widehat{\mathrm{MMD}}^2 = \frac{1}{n^2}\sum_{i,i'}k(x_i,x_{i'}) + \frac{1}{m^2}\sum_{\ell,\ell'}k(y_\ell,y_{\ell'}) - \frac{2}{nm}\sum_{i,\ell}k(x_i,y_\ell),
\end{equation}
where $k(x,y)=\exp(-\gamma\|x-y\|_2^2)$.
The bandwidth is fixed from reference pairs only, with $\gamma$ equal to one half divided by the median positive squared reference distance, and is shared across methods within a dataset.
MMD$^2$ uses $1000$ observations per sample for synthetic and wildfire data, and $512$ for HRRR.
These subsets are smaller than the full TCE evaluation samples.
MMD$^2$ assesses global distributional fit and complements the tail-specific curves; its bar plots use logarithmic vertical axes.

\subsection{CIFAR Rare-Label Recovery and Label TV}
\label{subsec:supp_cifar_metrics}

A balanced ResNet-18 classifier, with a $3\times3$ input convolution and no initial max pooling, assigns generated images to CIFAR classes.
Its training pool uses the original CIFAR training images while excluding the generative benchmark's validation and evaluation images.
The official CIFAR test images are split into two alternating halves: one selects classifier weights and checks the final acceptance precision, while the other calibrates the acceptance rule.
Classifier training uses SGD with learning rate $0.1$, momentum $0.9$, weight decay $5\times10^{-4}$, cosine scheduling, batch size $256$, and up to $80$ epochs with patience $12$. Training augmentation includes random crops and horizontal flips.

The rare set $\mathcal{R}$ contains the $20$ least frequent classes in the generative training split, with class index breaking frequency ties.
A rare prediction is accepted only if it meets a confidence threshold and a same-class feature-support threshold, using normalized classifier embeddings and a balanced real-image feature bank.
Support thresholds use the fifth percentile of calibration similarity within each rare class.
Confidence calibration targets at least $90\%$ weighted precision among accepted rare predictions.
Recovery is reported only when held-out rare-class accuracy is at least $60\%$, checked acceptance precision is at least $90\%$, and at least $30$ real rare predictions are accepted.

If $a_g$ and $a_r$ are accepted rare counts among $N_g$ generated and $N_r$ real evaluation images, rare recovery is
\begin{equation}
    \operatorname{Recovery}(\%)=100\,\frac{a_g/N_g}{a_r/N_r}.
\end{equation}
It is a ratio of accepted rare-label masses, not classifier accuracy. If $\widehat{h}_g(c)$ and $\widehat{h}_r(c)$ are predicted class frequencies, label TV in percentage points is
\begin{equation}
    \operatorname{TV}(\mathrm{pp})=50\sum_{c=1}^{100}|\widehat{h}_g(c)-\widehat{h}_r(c)|.
\end{equation}
The current evaluator has $75.1\%$ overall and $81.7\%$ rare-class accuracy on real evaluation images.
Its confidence cutoff is approximately $0.986$, checked acceptance precision is $90.2\%$, and $71$ of $1958$ real images receive accepted rare labels.
These checks support label-based comparisons but do not establish overall perceptual image quality.

\subsection{Tail-Quantile MSLE for the DLPM Comparison}
\label{subsec:supp_msle}

The additional DLPM experiment uses the tail-quantile MSLE associated with the protocol of \citet{shariatian2025heavy}. For absolute-coordinate quantile functions $Q^X_j$ and $Q^Y_j$, we compute
\begin{equation}
    \operatorname{MSLE}_{0.95}=\frac{1}{d}\sum_{j=1}^d
    \int_{0.95}^{1}\bigl[\log Q^X_j(u)-\log Q^Y_j(u)\bigr]^2\,\mathrm{d}u.
\end{equation}
The integral uses $1000$ midpoint quantile levels, with quantiles clipped below at $10^{-12}$ before taking logarithms.
Because it operates on coordinate magnitudes, this definition handles signed data.
It differs from pointwise squared differences of $\log(1+x)$ and is separate from the main TCE evaluation.

\section{Experimental Details}
\label{sec:supp_exp_details}

Experiments use single-GPU Slurm jobs with one NVIDIA V100 and eight CPU cores.
Each main training job runs the four trained models sequentially.
Epoch budgets below are maxima: early stopping can shorten training, and the best validation weights are restored.

\subsection{Datasets and Preprocessing}
\label{subsec:supp_datasets_detail}

\paragraph{Isotropic $\alpha$-stable.}
We generate centered symmetric isotropic $\alpha$-stable samples in $d=150$, with $\alpha=1.7$, using a Gaussian scale-mixture representation.
The fixed seeded split contains $25{,}000$ training, $1{,}000$ validation, and $10{,}000$ test-loss samples.
Distributional evaluation draws an independent reference sample of $5$M observations and $5$M generated observations per method.
No standardization is applied.

\paragraph{Unbalanced $\alpha$-stable mixture.}
This $150$-dimensional target has $16$ modes in a rank-$6$ latent subspace, power-law mode probabilities with imbalance parameter $\tau=1.2$, mean scale $7.5$, base scale $0.55$, anisotropy $1.0$, and stable local noise with $\alpha=1.7$.
The structure seed is zero. Splits, evaluation counts, and standardization follow the isotropic benchmark.

\paragraph{HRRR East and West.}
We use the NOAA HRRR analysis archive processed by dynamical.org~\citep{dowell2022hrrr,dynamical2026hrrr}.
Each hourly vector contains precipitation at $150$ fixed U.S. cities, selected east or west of longitude $-100^\circ$ and mapped to distinct nearest HRRR grid cells.
Precipitation rates are converted to mm/hour.
The requested period is October 6, 2014 through September 30, 2026; incomplete aligned hours are excluded.
Both regions contain $103{,}018$ complete hours.
We remove $2{,}473$ all-zero East vectors and $10{,}367$ all-zero West vectors, leaving $100{,}545$ and $92{,}651$ wet hours.
No positive threshold is imposed: tiny nonzero observations remain.
A seeded random split reserves $20{,}000$ training and $1{,}000$ validation hours per region.
All remaining wet hours are evaluated: $79{,}545$ East and $71{,}651$ West.

\paragraph{Wildfire size.}
We sample distinct positive-size U.S. fire records from the seventh edition of FPA-FOD, covering 1992--2024~\citep{short2026wildfire}.
Each observation is final fire size in acres. The seeded split contains $25{,}000$ training, $1{,}000$ validation, and $1$M evaluation observations.

\paragraph{CIFAR100-LT.}
We construct a class-imbalanced subset of the CIFAR-100 training images, with imbalance factor $10$ and image shape $3\times32\times32$~\citep{krizhevsky2009learning}.
Ten percent is reserved for each of validation and evaluation, yielding $15{,}657/1{,}958/1{,}958$ images.
Pilots use a fixed seeded subset of $4096$ training images; final training uses the full training split.
Evaluation pairs all $1{,}958$ real evaluation images with $1{,}958$ generated images per method.
The configured generation cap is $5000$, but evaluation never exceeds the available real reference count.
The separate classifier pools are described in \autoref{subsec:supp_cifar_metrics}.

HRRR and wildfire coordinates are standardized using training means and standard deviations only; zero standard deviations are replaced by one.
The inverse transformation is applied before distributional evaluation.
Synthetic coordinates remain unstandardized.
CIFAR pixels are mapped to $[-1,1]$. All main splits and final training runs use seed zero.

\subsection{Models and Architectures}
\label{subsec:supp_models_detail}

\autoref{tab:model_summary} lists DDPM-V~\citep{ho2020denoising}, DDIM-V~\citep{song2021denoising}, DLPM~\citep{shariatian2025heavy}, t-EDM, Linear Gaussian flow~\citep{lipman2023flowmatching,liu2023flowstraight}, and two zero-network controls.
DDPM-V and DLPM use velocity (V) and noise prediction, respectively.
DDIM-V and DDPM-V share the same model weights, follow the variance-preserving (VP) formulation.
t-EDM uses the Student-$t$ construction of \citet{pandey2025heavytailed} with EDM preconditioning~\citep{karras2022elucidating}.
Linear Gaussian flow uses $x_t=(1-t)x_0+t x_1$ and a learned velocity field.
The zero controls set every network parameter to zero while retaining the sampler, schedule, and preconditioning; they are not simply independent samples drawn directly from the initial prior.
DDIM uses the trained DDPM network, with $\eta=0$, without separate training.
Step counts do not imply identical numbers of network evaluations across solvers.

\begin{table}[htbp]
    \centering
    \small
    \resizebox{1.0\linewidth}{!}{%
    \begin{tabular}{llllr}
        \toprule
        Model & Noise / path & Weights & Sampler & Steps \\
        \midrule
        DDPM-V & Gaussian & trained & DDPM & 128 \\
        DDIM-V & Gaussian & shared DDPM & DDIM, $\eta=0$ & 128 \\
        DLPM & stable, $\alpha=1.8$ & trained & native DLPM & 128 \\
        t-EDM & Student-$t$, $\nu=2.2$ & trained & stochastic Heun & 64 \\
        Linear Gaussian flow & linear Gaussian & trained & Heun & 128 \\
        DLPM (zero) & stable, $\alpha=1.8$ & zeroed & native DLPM & 128 \\
        t-EDM (zero) & Student-$t$, $\nu=2.2$ & zeroed & stochastic Heun & 64 \\
        \bottomrule
    \end{tabular}
    }
    \vspace{1em}
    \caption{Main-model settings. DDPM and flow noise scales are selected by the pilot. t-EDM fixes $\sigma_{\min}=0.005$, $\sigma_{\max}=5$, and $\sigma_{\mathrm{data}}=1$.}
    \label{tab:model_summary}
\end{table}

\begin{table}[htbp]
    \centering
    \small
    \resizebox{1.0\linewidth}{!}{%
    \begin{tabularx}{\linewidth}{lX}
        \toprule
        Data & Shared backbone \\
        \midrule
        All vector experiments & Time-conditioned MLP: width $240$, depth $4$, time embedding dimension $128$, normalization enabled, dropout zero. The same architecture is used in pilots, final training, learning curves, and the DLPM implementation comparison. \\
        CIFAR100-LT & Transformer: two layers, four attention heads of dimension $32$ (hidden width $128$), patch size $4$, continuous timestep conditioning, GELU approximation, adaptive layer normalization, and dropout zero. \\
        \bottomrule
    \end{tabularx}
    }
    \vspace{1em}
    \caption{Neural-network architectures. Vector input/output dimension matches the dataset; image input/output shape is $3\times32\times32$.}
    \label{tab:backbone_summary}
\end{table}

\subsection{Pilot Selection and Final Training}
\label{subsec:supp_pilot_runs_detail}
\label{subsec:supp_main_runs_detail}

One optimization setting is selected per dataset and trained model, using the mean validation objective over pilot trials.
Model-specific objective values are not compared across methods.
The pilot budget reported here is $16$ epochs, with three trials per candidate for vector data and one for CIFAR.
Vector pilots use the full training split; CIFAR pilots use $4096$ images.
The pilot and final stages use the same backbone.
\begin{table}[htbp]
    \centering
    \small
    \resizebox{0.78\linewidth}{!}{%
    \begin{tabular}{llll}
        \toprule
        Data & Learning-rate grid & $\sigma_{\max}$ grid & Candidates / trials \\
        \midrule
        Synthetic & $\{10^{-4},2\!\times\!10^{-4},5\!\times\!10^{-4}\}$ & $\{2,5,8\}$ & $9/3$ or $3/3$ \\
        HRRR, wildfire & $\{2\!\times\!10^{-4},5\!\times\!10^{-4}\}$ & $\{2,5\}$ & $4/3$ or $2/3$ \\
        CIFAR100-LT & $\{5\!\times\!10^{-5},10^{-4},2\!\times\!10^{-4}\}$ & $\{2,5,8\}$ & $9/1$ or $3/1$ \\
        \bottomrule
    \end{tabular}
    }
    \vspace{1em}
    \caption{Pilot grids. DDPM-V and Linear Gaussian flow tune learning rate and $\sigma_{\max}$; DLPM and t-EDM tune learning rate only. Candidate/trial counts are given in that order.}
    \label{tab:pilot_summary}
\end{table}

\begin{table}[htbp]
    \centering
    \small
    \resizebox{0.82\linewidth}{!}{%
    \begin{tabular}{lrrrrrr}
        \toprule
        Data & Batch & Epochs & Patience & Weight decay & Warmup & Clip norm \\
        \midrule
        Synthetic & 1024 & 720 & 100 & $10^{-6}$ & 400 & 10 \\
        HRRR, wildfire & 512 & 720 & 100 & $10^{-6}$ & 100 & 10 \\
        CIFAR100-LT & 256 & 720 & 24 & $10^{-4}$ & 50 & 5 \\
        \bottomrule
    \end{tabular}
    }
    \vspace{1em}
    \caption{Final training settings. Epochs are maxima, patience is in epochs, and warmup is in optimizer updates. AdamW uses the pilot-selected learning rate and a cosine schedule ending at $5\%$ of that rate. Each main model is trained once.}
    \label{tab:training_summary}
\end{table}

Pilot early-stopping patience is $32$ epochs for vector data and $4$ for CIFAR, with $16$ warmup updates for CIFAR pilots.
The synthetic $10$k held-out split is used for test loss, not as the $5$M distributional reference.

\section{Additional Empirical Results}
\label{sec:supp_exp_addi}

\subsection{Additional Vector-Distribution Results}
\label{subsec:supp_vector_results}

\begin{figure}[htbp]
    \centering
    \begin{minipage}{0.4\linewidth}
        \centering
        \includegraphics[width=\linewidth,trim=0 84bp 0 0,clip]{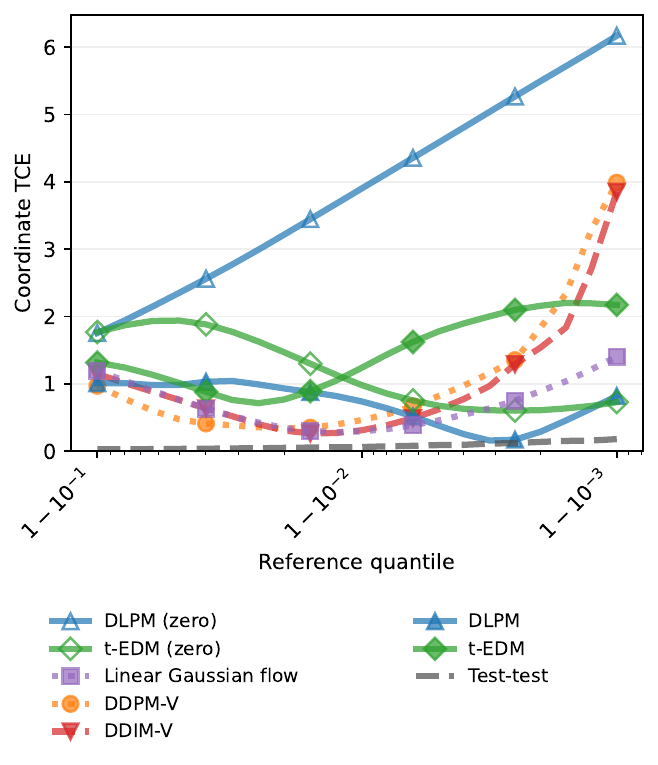}
    \end{minipage}

    \includegraphics[width=0.58\linewidth,trim=15bp 6bp 35bp 289bp,clip]{figures/synthetic_heavy_tail__coordinate_tce__curve.pdf}
    \caption{Coordinate TCE on HRRR West, using the same conventions as \autoref{fig:main_coordinate_tce}.}
    \label{fig:supp_hrrr_west}
\end{figure}

\begin{figure}[p]
    \centering
    \begin{tabular}{@{}cc@{}}
        \includegraphics[width=0.3\linewidth]{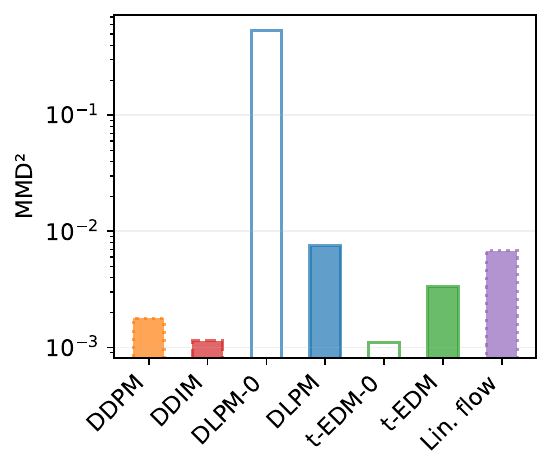} &
        \includegraphics[width=0.3\linewidth]{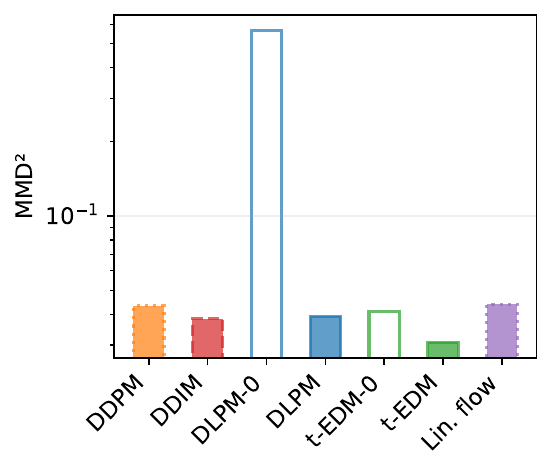} \\
        \small (a) Isotropic $\alpha$-stable & \small (b) Unbalanced $\alpha$-stable mixture \\[0.6em]
        \includegraphics[width=0.3\linewidth]{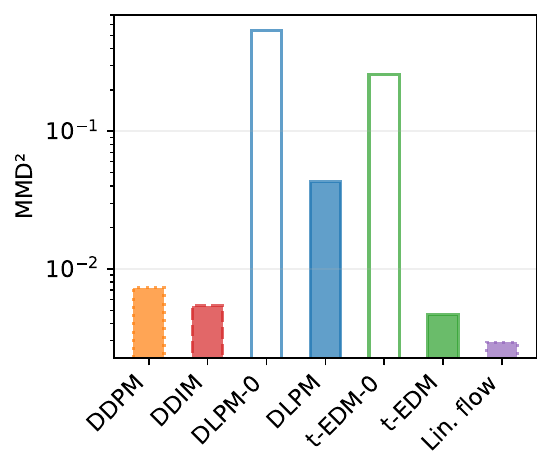} &
        \includegraphics[width=0.3\linewidth]{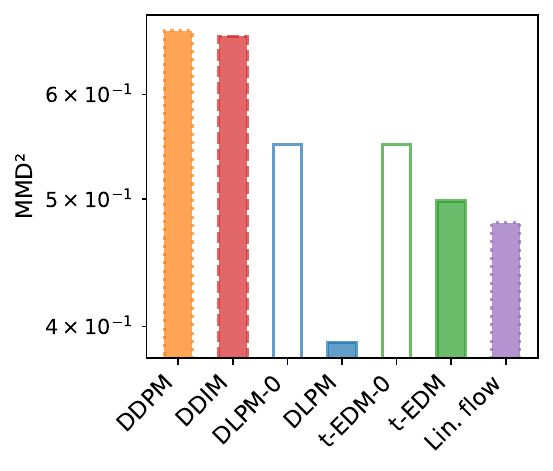} \\
        \small (c) HRRR West & \small (d) Wildfire size
    \end{tabular}
    \caption{Remaining MMD$^2$ results. Lower is better; vertical axes are logarithmic. The suffix ``0'' denotes a zero-network control.}
    \label{fig:supp_remaining_mmd}
\end{figure}

\autoref{fig:supp_hrrr_west} completes the coordinate-TCE comparison with HRRR West.
The remaining MMD$^2$ bar plots are collected in \autoref{fig:supp_remaining_mmd}.

\subsection{Sample-Size Dependence}
\label{subsec:supp_learning_curves}

We compare DDPM-V and DLPM on isotropic $\alpha$-stable targets with data tail index $1.7$, in dimensions $d\in\{5,30,50\}$.
We report training sizes $n\in\{10,100,1000,10000\}$, averaged over five trials.
Within each dimension/trial pair, training subsets are nested, and validation and reference pools use independent random streams.
Validation uses $1000$ samples and distributional evaluation generates $100{,}000$ samples per model.
All methods use the common MLP of \autoref{tab:backbone_summary}; DLPM fixes noise index $1.8$, while DDPM-V uses $\sigma_{\max}=5$. Sampling uses $128$ steps.

Each trained model receives exactly $10{,}000$ optimizer updates, with replacement minibatches of size $1024$, rather than a fixed epoch count.
AdamW uses learning rate $5\times10^{-4}$, weight decay $10^{-6}$, gradient clipping at norm $10$, $400$ warmup updates, and cosine decay to $5\%$ of the initial rate.
The initialization and minibatch-index streams are reset for paired comparisons across training sizes.
We evaluate global fit using MMD$^2$ on $1000$ reference and $1000$ generated observations, following \autoref{subsec:supp_mmd}. \autoref{tab:learning_curves} reports the five-trial means.

\begin{table}[htbp]
    \centering
    \small
    \resizebox{1.0\linewidth}{!}{%
    \begin{tabular}{lrrrrrr}
        \toprule
        Train & \multicolumn{2}{c}{$d=5$} & \multicolumn{2}{c}{$d=30$} & \multicolumn{2}{c}{$d=50$} \\
        \cmidrule(lr){2-3}\cmidrule(lr){4-5}\cmidrule(lr){6-7}
        & DDPM-V & DLPM & DDPM-V & DLPM & DDPM-V & DLPM \\
        \midrule
        10 & $5.850\times10^{-2}$ & $3.119\times10^{-1}$ & $1.355\times10^{-1}$ & $2.812\times10^{-1}$ & $9.815\times10^{-2}$ & $2.871\times10^{-1}$ \\
        100 & $1.464\times10^{-2}$ & $1.833\times10^{-1}$ & $7.386\times10^{-2}$ & $2.638\times10^{-1}$ & $9.168\times10^{-2}$ & $2.458\times10^{-1}$ \\
        1k & $1.459\times10^{-3}$ & $2.136\times10^{-3}$ & $2.406\times10^{-2}$ & $2.604\times10^{-1}$ & $5.065\times10^{-2}$ & $2.537\times10^{-1}$ \\
        10k & $2.275\times10^{-3}$ & $6.656\times10^{-4}$ & $2.308\times10^{-3}$ & $1.033\times10^{-3}$ & $1.142\times10^{-3}$ & $1.984\times10^{-2}$ \\
        \bottomrule
    \end{tabular}%
    }
    \vspace{1em}
    \caption{Mean MMD$^2$ over five trials versus training size, for $d\in\{5,30,50\}$. Lower is better.}
    \label{tab:learning_curves}
\end{table}

DDPM-V achieves lower mean MMD$^2$ across all dimensions for training sizes up to $n=1000$, suggesting that DLPM faces greater statistical learning difficulties in data-scarce regimes.
However, at $n=10000$, DLPM outperforms DDPM-V for $d=5$ and $d=30$, while DDPM-V remains better for $d=50$.
This highlights the role of task difficulty: DLPM becomes more competitive as the sample size increases and the dimension decreases.
These observations support our theoretical intuition that DLPM can face additional learning challenges, which may be mitigated by sufficient data or lower dimensionality.

\subsection{DLPM empirical claim reproduction}
\label{subsec:supp_dlpm_reproduction}

Additionally, we reproduce the experimental setting of Table 1 in \citet{shariatian2025heavy}, which reports improved tail coverage and robustness for DLPM on heavy-tailed data.
To assess reproducibility, we compare our independent DLPM implementation, a thin wrapper around the authors' original implementation, and their published results, using the MSLE metric adopted in the original study.
Despite these complementary implementations, we are unable to recover the reported performance ranking.
As shown in~\autoref{tab:shariatan-et-al-msle}, this discrepancy raises questions about the reproducibility of the original empirical findings.

\begin{table}[ht]
    \centering
    \small
    \setlength{\tabcolsep}{4pt}
    \resizebox{\linewidth}{!}{%
        \begin{tabular}{lcccc}
            \toprule
            $\mathrm{MSLE}_{0.95}$ & $\alpha = 1.5$ & $\alpha = 1.7$ & $\alpha = 1.9$ & $\alpha = 2.0$ \\
            \midrule
            DLPM (our code) & $0.203 \pm 0.386$ & $0.002 \pm 0.001$ & $0.004 \pm 0.003$ & $0.007 \pm 0.009$ \\
            DLPM (author code wrapped) & $0.079 \pm 0.071$ & $0.005 \pm 0.008$ & $0.003 \pm 0.002$ & $0.009 \pm 0.007$ \\
            DLPM (paper results) & $0.160 \pm 0.128$ & $0.071 \pm 0.028$ & $0.132 \pm 0.101$ & $0.798 \pm 0.601$ \\
            \bottomrule
        \end{tabular}%
    }
    \vspace{1em}
    \caption{Shariatian et al. (2025) protocol comparing our DLPM and the wrapped author implementation. Lower $\mathrm{MSLE}_{0.95}$ is better.}
    \label{tab:shariatan-et-al-msle}
\end{table}

\section{Additional Remarks and Discussion}
\label{sec:supp_additional_remarks}

We collect here additional remarks and discussions that provide further insight into our analysis.

\subsection{Structural lower bounds.}
\label{subsec:supp_lower_bound}

As for VE-GD, each component of \autoref{eq:DLPM_sampling_bound} is unavoidable on an appropriate subclass of problems.
\emph{\underline{Initialization:}}
Writing $Y_T=b_T(S_\alpha+\rho_T Y_0)$, a deterministic translated initial condition $Y_0=y$ gives $\mathrm{TV}\big(\mathcal L(Y_T),\mathcal L(b_TS_\alpha)\big)\gtrsim|\rho_T|\,\normEc{y}$ for $|\rho_T|$ small, which is exponentially small in $T$ when $|\rho_T|\asymp e^{-cT}$. As for VE-GD, this term is independent of $n$ and is not a statistical error.
\emph{\underline{Approximation:}}
Classical approximation theory for Hölder classes yields the lower bound $m^{-\beta(\alpha)/d}$ for the local approximation error of the reverse object \citep{Tsybakov2009}.
\emph{\underline{Estimation:}}
Assouad and Fano arguments give the classical minimax rate $n^{-\beta(\alpha)/(2\beta(\alpha)+d)}$ for estimating the reverse object \citep{Tsybakov2009}.
It remains to propagate these local errors to the generated distribution.
The factor $T$ in \autoref{eq:DLPM_sampling_bound} enters through the telescoping decomposition of the reverse chain (see \autoref{subsec:supp_DLPM_proof}), which bounds the global error by the sum of the $T$ local kernel errors.
The next proposition shows that this linear accumulation cannot be improved in general.

\begin{proposition}[Tightness of the $T$-accumulation]
    \label{prop:lower_bound_T}
    Let $\varepsilon\in(0,1)$ and $T\ge1$.
    There exist a measurable space $\Omega$, an initial distribution $\mu_T$ on $\Omega$, true reverse Markov kernels $(Q_t)_{t=1}^T$ and learned kernels $(\widehat Q_t)_{t=1}^T$ on $\Omega$ such that
    \begin{equation*}
        \sup_{x\in\Omega} \mathrm{TV}\big(Q_t(x,\cdot),\widehat Q_t(x,\cdot)\big)\le\varepsilon \quad\text{for all } t\in\{1,\dots,T\}\eqsp,
    \end{equation*}
    and
    \begin{equation*}
        \mathrm{TV}\big(\mu_TQ_T\cdots Q_1,\ \mu_T\widehat Q_T\cdots\widehat Q_1\big) = 1-(1-\varepsilon)^T \ge \tfrac12\min\{1,T\varepsilon\}\eqsp.
    \end{equation*}
\end{proposition}
The proof, given in \autoref{subsec:proof_prop_T}, introduces at each reverse step a single ``error channel'' of mass $\varepsilon$.
Each channel is activated exactly once along the trajectory and all errors push mass in the same direction, so they accumulate without cancellation.
Since the telescoping argument gives the matching upper bound $T\varepsilon$, the linear dependence on $T$ is sharp up to a universal constant.
Taking $\varepsilon\asymp m^{-\beta(\alpha)/d}$ (approximation) or $\varepsilon\asymp\sqrt{\mathrm{Comp}/n}$ (estimation) recovers the $T$-linear scaling of \autoref{thm:error-DLPM}: without additional contractivity or cancellation properties of the reverse chain, the factor $T$ is intrinsic to the reverse-chain structure rather than a proof artifact.
As for VE-GD, these lower bounds hold term by term and do not constitute a complete minimax lower bound for the sum.

\subsection{What the comparison does and does not say.}
\label{subsec:supp_comparison_discussion}

At first sight, the DLPM bound appears more favorable, in agreement with the intuitive principle that heavy-tailed data should be generated using heavy-tailed noise: its initialization error decays exponentially fast, as $e^{-cT}$, whereas the VE-GD initialization error only decays polynomially, as $T^{-1/2}$. This exponential rate is a genuine strength of DLPM; in our analysis, it follows from \autoref{ass:exp_rho_decay} on the noise schedule. Note that the comparison below concerns two representative formulations, Gaussian VE diffusion and DLPM, and not light-tailed versus heavy-tailed diffusion models in general.
These formulations also differ in their time structure.
\autoref{thm:error-DDPM} is stated for the continuous-time VE dynamics \autoref{eq:general-fwd-dynamics} and its time reversal, so that the bound contains no time-discretization error and $T$ is a diffusion horizon.
\autoref{thm:error-DLPM} is stated for the discrete-time DLPM chain \autoref{eq:DLPM_Markov_kernel}, made of $T$ reverse Markov kernels, so that $T$ is both the horizon and the number of reverse steps; the linear factor $T$ in its training error counts the local kernel errors accumulated along the chain.

\autoref{thm:error-DDPM} and \autoref{thm:error-DLPM} provide a common framework for comparing the sampling guarantees of the two models through the same \emph{initialization-training} decomposition.
This shared structure makes it possible to identify how model-specific terms contribute to the error bounds, although it does not establish an ordering of the actual sampling errors.
The analyses account for the distinct objects learned by each model and retain the regularity assumptions of their original formulations, making explicit the conditions under which each guarantee holds.
On one hand, VE-GD learns the score $\nabla\log p_t$ of Gaussian-smoothed marginals, so that Sobolev regularity of $\pidata$ is the natural assumption, Gaussian convolution transferring it to the smoothed marginals.
On the other hand, DLPM learns the conditional reverse kernels of the $\alpha$-stable chain; since no quantitative transfer from a Sobolev assumption on $\pidata$ to this reverse object is known, we impose a standard Hölder assumption directly on it.
The two assumptions play parallel methodological roles, each tailored to the object learned by its respective model.
Together with the structural lower bounds above, which show that each term is attained on a suitable subclass of problems, these results provide a principled basis for comparing the mechanisms governing the two error bounds. This yields the following \emph{mechanistic} conclusion:
\begin{center}
    \emph{DLPM buys faster initialization at the price of a training cost that grows linearly with $T$.}
\end{center}
Since a large $T$ is precisely what enables the initialization gain, this cost may dominate in finite-sample regimes.

\subsection{Understanding DLPM Training Instability}
\label{subsec:supp_instability}

In our experiments, we observe that DLPM can exhibit greater run-to-run variability.
Several factors may contribute to this behavior, including heavy-tailed perturbations, potentially more variable stochastic gradients, and the difficulty of estimating reverse dynamics in low-density regions.
Our theoretical analysis also shows that local approximation and estimation errors can accumulate along the reverse chain, potentially amplifying these difficulties.
However, our experiments do not isolate the respective contributions of these mechanisms.
A systematic investigation of these mechanisms remains an interesting direction for future work.

\end{document}